\documentclass[11pt,letterpaper]{article}

\usepackage[T1]{fontenc}
\usepackage[utf8]{inputenc}
\usepackage{lmodern}
\usepackage[margin=0.95in,headheight=14pt,headsep=16pt,footskip=26pt]{geometry}
\usepackage{amsmath,amssymb}
\usepackage{graphicx}
\usepackage{booktabs,tabularx,multirow,array}
\usepackage{microtype,caption,fancyhdr,placeins,enumitem}
\usepackage{xcolor}
\usepackage[numbers,sort&compress]{natbib}

\definecolor{linkblue}{RGB}{24,58,98}
\PassOptionsToPackage{hyphens}{url}
\usepackage[colorlinks=true,allcolors=linkblue,bookmarksopen=true]{hyperref}
\expandafter\def\expandafter\UrlBreaks\expandafter{\UrlBreaks\do\_\do\-}
\newsavebox{\tablebox}
\newcommand{\fittable}[1]{%
  \sbox{\tablebox}{#1}%
  \ifdim\wd\tablebox>\linewidth
    \resizebox{\linewidth}{!}{\usebox{\tablebox}}%
  \else
    \usebox{\tablebox}%
  \fi
}
\newcommand{\ci}[2]{\shortstack[r]{#1\\[-1pt]{\scriptsize #2}}}

\fancypagestyle{plain}{\fancyhf{}\fancyfoot[C]{\thepage}}
\AtBeginDocument{\thispagestyle{plain}}

\hypersetup{
  pdftitle={Cross-Precision Transfer, Non-Local Context, and Capacity--Readout Boundaries of Scale Calibration in Low-Bit LLMs},
  pdfauthor={Xu Cao and Wenxuan Xiao}
}

\begin{document}

\begin{center}
  {\LARGE\bfseries Cross-Precision Transfer, Non-Local Context, and\\[3pt]
  Capacity--Readout Boundaries of Scale Calibration in Low-Bit LLMs\par}
  \vspace{9pt}
  {\large\bfseries Xu Cao and Wenxuan Xiao\par}
  \vspace{2pt}
  {\small Astrmira Tech.\par}
  {\small \href{mailto:x.cao@astrmira.com}{\texttt{x.cao@astrmira.com}}, \href{mailto:w.xiao@astrmira.com}{\texttt{w.xiao@astrmira.com}}\par}
\end{center}
\vspace{2pt}

\begin{abstract}
Low-bit post-training quantization (PTQ) freezes discrete integer codes and zero points, leaving continuous groupwise scales as a compact adaptation interface. Through controlled interventions on Qwen3-1.7B and Llama-3.2-3B under composite language-modeling and distillation objectives, we test when learned relative scale gains transfer across representations, how surrounding network context shapes them, and where optimizer capacity and evaluation prompts govern low-bit behavior. First, installing relative gains ($s_g \mapsto a_g s_g$) bidirectionally between floating-point ($F$) and asymmetric INT3 ($Q$) Qwen3-1.7B models with zero recipient training retains $76.9\%$--$90.5\%$ of self-calibration negative log-likelihood (NLL) reductions on fresh unseen WikiText and C4 text alongside resolved own-source preferences; group permutation destroys transfer, norm-matched single-layer withdrawal degrades all $16/16$ layer--domain cells, and a Llama-3.2-3B control with bit-identical initial tensors at all calibrated projections isolates non-local context preferences. Second, expanding Llama-3.2-3B calibration from $4$ projections ($0.79\text{M}$ scales) to all $196$ projections ($22.02\text{M}$ scales) makes learning rate $0.01$ win a $50$-step training-loss pilot; continuing to $200$ steps memorizes the $28$-block WikiText pool, inflating unseen WikiText perplexity $2.06\times$--$2.28\times$ over $4$-site calibration and degrading mixed-action PIQA accuracy, whereas step size $0.001$ prevents memorization and restores PIQA accuracy. Third, across $24$ Qwen3-1.7B ARC readout endpoints, advancing from $200$ to $600$ steps worsens INT3 teacher KL on plain-text readouts yet reverses sign under native chat wrapping ($-0.287\text{ nats}$). Cross-fitted $4$-parameter static option-bias estimation explains $71.4\%$--$75.4\%$ of quantized $\text{Chat}-\text{Space}$ score-shift SSE ($<2.3\%$ on floating point); subtracting this bias lifts INT3 chat teacher agreement from $33.3\%$--$33.9\%$ to $56.6\%$--$57.5\%$ and cuts KL from $4.15$--$4.42$ to $1.18$--$1.24\text{ nats}$.
\end{abstract}

\section{Introduction}
\label{sec:intro}

Sub-$4$-bit post-training quantization (PTQ) compresses large language models (LLMs) by partitioning each linear weight matrix into small input-channel groups parameterized by frozen integer codes, zero points, and floating-point step sizes \citep{nagel2020adaround,li2021brecq,frantar2023gptq,xiao2023smoothquant,lin2024awq}. Once discrete codes are fixed, optimizing continuous groupwise scales substantially improves language-modeling and task losses while preserving low-bit storage and matrix-multiplication kernels---a property exploited both for post-quantization step-size refinement \citep{shao2024omniquant,chen2025efficientqat,kim2025guidedquant} and for parameter-efficient scale-only fine-tuning \citep{kim2023peqa}.

Despite widespread adoption, three practical and mechanistic questions remain open when scales are calibrated on composite objectives combining text NLL with teacher-referenced option KL. First, because a quantized checkpoint differs from its floating-point parent both at the calibrated projections and across surrounding quantized layers, it is unclear how much of a learned scale update's loss reduction can be transplanted directly across precisions or across distinct quantization configurations without re-optimization---and whether recipient-specific preferences still arise when local initial weights at the calibrated sites are identical. Second, while small-pool calibration overfitting is a classic concern in weight-reconstruction PTQ \citep{hubara2021accurate}, its behavior when unlocking all $>20\text{M}$ group scales of a $3\text{B}$ LLM under log-scale Adam updates---particularly when hyperparameters are screened on consumed calibration loss---has not been cleanly characterized. Third, low-bit models exhibit confidence distortions \citep{proskurina2024quantization} and prompt/label prior biases \citep{zhao2021calibrate}; how identical scale-calibrated checkpoints behave across downstream scoring normalizations and chat-template wrappers remains poorly understood.

\begin{table}[!t]
\centering
\caption{Overview of controlled interventions across $14$ unit-level cohorts and a $24$-endpoint ARC readout audit on Qwen3-1.7B and Llama-3.2-3B. All intervals use $10{,}000$ paired bootstrap resamples with zero post-hoc checkpoint selection.}
\label{tab:overview}
\setlength{\tabcolsep}{5pt}
\renewcommand{\arraystretch}{1.16}
\small
\begin{tabularx}{\linewidth}{@{}>{\raggedright\arraybackslash}p{0.10\linewidth}>{\raggedright\arraybackslash}p{0.22\linewidth}>{\raggedright\arraybackslash}X>{\raggedright\arraybackslash}X@{}}
\toprule
Part & Models \& Scope & Controlled Intervention & Principal Empirical Finding \\
\midrule
\textbf{I} (\S\ref{sec:setup}--\ref{sec:act1}) & Qwen3-1.7B ($F \leftrightarrow Q$); Llama-3.2-3B ($S \leftrightarrow J$) & Bidirectional $F \leftrightarrow Q$ installation; within-layer group shuffle; matched-$\|\theta\|_2$ layer drop; identical-local-tensor $S \leftrightarrow J$ swap & Cross-precision transfer retains $76.9\%$--$90.5\%$ of NLL benefit with resolved own-source preference; group shuffle destroys transfer; identical local tensors isolate non-local context preference ($98.6\%$--$99.3\%$ retained). \\
\addlinespace[3pt]
\textbf{II} (\S\ref{sec:act2}) & Llama-3.2-3B INT3 ($Q_0, S, J$; $12$ states) & $4$-site ($0.79\text{M}$) vs.\ $196$-site ($22.02\text{M}$) scales $\times$ $\text{lr} \in \{10^{-2}, 10^{-3}\}$ $\times$ $\{50, 200\}$ steps & A $50$-step FIT pilot selects $\text{lr}=10^{-2}$; $200$ steps memorize $28$ Wiki blocks ($2.06\times$--$2.28\times$ val PPL vs.\ $4$-site; $-1.90$ to $-2.56\text{ pp}$ PIQA on $S/J$); $\text{lr}=10^{-3}$ avoids memorization and recovers $+2.67$ to $+2.83\text{ pp}$ PIQA. \\
\addlinespace[3pt]
\textbf{III} (\S\ref{sec:act3}) & Qwen3-1.7B ($24$ ARC endpoints; PIQA) \& Llama-3.2-3B (PIQA) & PIQA length normalization; $4$ ARC interfaces across $0 \to 200 \to 600$ steps; out-of-fold bias subtraction & $200 \to 600$ INT3 ARC KL worsens on plain prompts ($+0.049$ to $+0.082$) yet flips to $-0.287$ in chat; subtracting $5$-fold static option bias lifts INT3 chat teacher agreement from $33\%$--$34\%$ to $57\%$--$58\%$. \\
\bottomrule
\end{tabularx}
\end{table}

We address these three questions through controlled interventions on Qwen3-1.7B \citep{qwen2025qwen3,qwen2025modelcard} and Llama-3.2-3B \citep{meta2024llama32}, summarized in Table~\ref{tab:overview}:
\begin{enumerate}[leftmargin=1.6em,itemsep=2pt,topsep=2pt]
  \item \textbf{Cross-precision benefit transfer and non-local context controls (\S\ref{sec:setup}--\ref{sec:act1}).} Expressing updates as dimensionless groupwise multipliers ($s_g \mapsto a_g s_g$) enables exact zero-training installation between floating-point ($F$) and GPTQ INT3 ($Q$) checkpoints. On strictly isolated fresh unseen Qwen3-1.7B WikiText and C4 cohorts, bidirectional cross-installation retains $76.9\%$--$90.5\%$ of self-calibration NLL reductions while exhibiting reciprocal own-source preferences; a four-term crossed accounting identity explains why $a_F$ cuts INT3 C4 perplexity by $21.6\%$ while simultaneously widening the $Q$--$F$ C4 loss gap ($0.2774 \to 0.3017\text{ nats}$). Within-projection group permutation destroys transfer, norm-matched single-layer withdrawal degrades all $16/16$ domain--layer cells, and a Llama-3.2-3B INT3 control ($S$ vs.\ $J$) with bit-for-bit identical initial tensors at all four calibrated projections demonstrates statistically resolved own-context preferences induced purely by nine uncalibrated projections elsewhere (across all C4 block-bootstrap lengths $L \in \{1, 2, 4, 8, 16\}$).
  \item \textbf{Small-pool capacity overfitting and step-size control (\S\ref{sec:act2}).} Expanding trainable scales on Llama-3.2-3B from $4$ projections ($0.79\text{M}$) to all $196$ projections ($22.02\text{M}$) on a fixed small PTQ pool ($28$ WikiText blocks, $256$ C4 blocks, $592$ ARC items) exposes a consumed-loss pilot trap: $\text{lr}=0.01$ achieves lower $50$-step consumed calibration loss than $\text{lr}=0.001$ ($2.103$ vs.\ $2.373$). Continuing to $200$ steps memorizes the $28$ WikiText blocks (mean sampled-batch NLL of $0.17$--$0.29\text{ nats}$ over steps $181$--$200$), inflating unseen validation WikiText perplexity $2.06\times$--$2.28\times$ relative to $4$-site calibration and dropping PIQA accuracy by $1.90$--$2.56\text{ pp}$ on $S/J$. Step-size control ($\text{lr}=0.001$) prevents memorization, recovers $+2.67$ to $+2.83\text{ pp}$ on PIQA, and reveals asynchronous $50 \to 200$ step trajectories across ARC, C4, and WikiText.
  \item \textbf{Readout sign reversal and cross-fitted option-bias subtraction (\S\ref{sec:act3}).} Across downstream PIQA and a zero-training $24$-endpoint Qwen3-1.7B ARC readout audit ($569$ items $\times$ $6$ checkpoints $\times$ $4$ interfaces), proxy-loss improvements decouple from discrete decisions. Advancing Qwen INT3 from $200$ to $600$ updates worsens teacher-referenced option KL under bare ($+0.050$), space-prefixed ($+0.049$, despite $99.97\%$ single-token teacher mass), and full-text ($+0.082$) interfaces, yet flips sign to $-0.287\text{ nats}$ under the native chat wrapper ($-0.338\text{ nats}$ difference-in-differences vs.\ bare). Symmetric bilinear KL accounting isolates student score shifts ($-0.352\text{ nats}$) as the dominant term, and $5$-fold cross-fitting on $546$ four-choice items shows that static option-label/position bias accounts for $71.4\%$--$75.4\%$ of quantized $\text{Chat}-\text{Space}$ centered shift SSE ($76.8\%$--$82.8\%$ on same-token $\text{Chat}-\text{Bare}$) versus $<2.3\%$ on floating point; subtracting this out-of-fold $4$-vector bias directly raises held-out INT3 chat teacher agreement from $33.3\%$--$33.9\%$ to $56.6\%$--$57.5\%$.
\end{enumerate}

\section{Relative Scale Interface and Crossed Accounting}
\label{sec:setup}

\subsection{Groupwise Affine Coordinates and Native Scale Absorption}
\label{sec:setup_coords}

Consider a linear projection $W \in \mathbb{R}^{d_{\mathrm{out}} \times d_{\mathrm{in}}}$ partitioned along its input dimension into $G = d_{\mathrm{in}} / g_{\mathrm{size}}$ contiguous channel groups ($g_{\mathrm{size}} = 128$). Each output channel $o \in \{1,\dots,d_{\mathrm{out}}\}$ and input group $c \in \{1,\dots,G\}$ defines a weight slice $w_g \in \mathbb{R}^{g_{\mathrm{size}}}$ indexed by $g = (o, c)$. In an asymmetric $3$-bit integer encoding ($Q$), dequantized group weights take the affine form
\begin{equation}
  w_{Q,g,0} = s_{g,0}(q_g - z_g),
  \label{eq:affine_group}
\end{equation}
with frozen unsigned integer codes $q_g \in \{0,\dots,7\}^{g_{\mathrm{size}}}$, scalar zero point $z_g \in \mathbb{R}$, and initial positive group scale $s_{g,0} > 0$; on the aligned floating-point model ($F$), the corresponding slice is $w_{F,g,0} \in \mathbb{R}^{g_{\mathrm{size}}}$. Parameterizing calibration over $g \in \mathcal{G}_{\mathrm{cal}}$ via dimensionless positive multipliers $a_g = \exp(\theta_g) > 0$ initialized at $\theta_g = 0$ ($a_g = 1$) acts homogeneously on both representations:
\begin{equation}
  w_{F,g}(a_g) = a_g w_{F,g,0}, \qquad
  w_{Q,g}(a_g) = a_g w_{Q,g,0} = (a_g s_{g,0})(q_g - z_g).
  \label{eq:scale_absorption}
\end{equation}
From a calibrated floating-point donor, $a_g$ is read directly from the saved multiplier tensor; from a quantized donor with terminal scale $s_{g,\star} > 0$, it is extracted as $a_g = s_{g,\star} / s_{g,0}$. Installing any donor gain vector $a \in \mathbb{R}_{>0}^{|\mathcal{G}_{\mathrm{cal}}|}$ onto a native quantized recipient updates only its scale slots $s_g \leftarrow a_g s_{g,0}$ without altering $q_g$, $z_g$, memory layout, or low-bit matrix-multiplication kernels. Our installer promotes donor and recipient scales to IEEE-754 FP64, multiplies in FP64, verifies positivity, finiteness, and canonical input-column group alignment (mapping GPTQ's \texttt{act\_order} permutation back to original input columns), and casts once to FP32 upon writeback, reproducing donor checkpoints bit-for-bit under self-installation ($3.27\text{ ms}$ native scale arithmetic on CPU for Qwen's $393{,}216$ groups; full I/O breakdown in Appendix~\ref{app:protocol}, Table~\ref{tab:a_cpu}).

\subsection{Local Displacement and Four-Term Crossed Accounting}
\label{sec:setup_accounting}

Writing $\delta_g := w_{Q,g,0} - w_{F,g,0}$ for the initial quantization residual of group $g$ and $\Delta w_{B,g}(a_g) := w_{B,g}(a_g) - w_{B,g,0}$ for the additive weight change on base $B \in \{F, Q\}$, distributivity immediately gives
\begin{equation}
  \Delta w_{Q,g}(a_g) - \Delta w_{F,g}(a_g) = (a_g - 1)(w_{Q,g,0} - w_{F,g,0}) = (a_g - 1)\delta_g.
  \label{eq:local_displacement}
\end{equation}
Unlike additive task vectors \citep{ilharco2023editing}, applying a common relative gain $a_g$ induces additive weight displacements on $Q$ and $F$ that differ locally by $(a_g - 1)\delta_g$, alongside static quantization residuals at uncalibrated layers.

To account for shared benefit versus recipient-specific preference on any per-unit loss $L(B, a)$ across bases $B \in \{F, Q\}$ and gains $a \in \{\mathbf{1}, a_F, a_Q\}$, define each recipient's \emph{quantized-minus-floating source difference} $d_B$ and the \emph{common-gain precision gap} $G(a)$:
\begin{equation}
  d_B := L(B, a_Q) - L(B, a_F) \quad (B \in \{F, Q\}), \qquad
  G(a) := L(Q, a) - L(F, a).
  \label{eq:defs_crossed}
\end{equation}
On losses where lower is better, $Q$ prefers its own gain $a_Q$ when $d_Q < 0$, and $F$ prefers its own gain $a_F$ when $d_F > 0$. Regrouping the four crossed states yields two exact unit-level accounting identities:
\begin{align}
  \mathcal{I} \;:=\; d_Q - d_F
  &\;=\; \bigl[L(Q, a_Q) - L(Q, a_F)\bigr] - \bigl[L(F, a_Q) - L(F, a_F)\bigr]
   \;=\; G(a_Q) - G(a_F), \label{eq:crossed_identity} \\
  G(a_F) - G(\mathbf{1})
  &\;=\; \bigl[L(Q, a_F) - L(Q, \mathbf{1})\bigr] - \bigl[L(F, a_F) - L(F, \mathbf{1})\bigr]. \label{eq:gap_paradox}
\end{align}
Equation~\eqref{eq:gap_paradox} shows why evaluating low-bit calibration solely via the precision gap $G(a)$ can mislead: if $a_F$ substantially lowers loss on both bases ($L(Q, a_F) \ll L(Q, \mathbf{1})$ and $L(F, a_F) \ll L(F, \mathbf{1})$) but reduces NLL slightly more on clean $F$ than on quantized $Q$, then $G(a_F) > G(\mathbf{1})$ even though $Q$ improves substantially. Alongside paired loss differences, we summarize recipient $B$'s descriptive \emph{retained self-calibration benefit} under cross-installed gain $a_{B'}$ ($B' \neq B$) as
\begin{equation}
  R_B := \frac{L(B, \mathbf{1}) - L(B, a_{B'})}{L(B, \mathbf{1}) - L(B, a_B)},
  \label{eq:retained_ratio}
\end{equation}
which measures the fraction of recipient self-calibration loss reduction achieved without recipient optimization; because neural losses are nonlinear in parameters, high $R_B$ establishes functional benefit portability rather than an orthogonal decomposition of learned parameter vectors.

\section{Cross-Precision Transfer and Non-Local Network Context}
\label{sec:act1}

\subsection{Bidirectional FP--INT3 Gain Transfer on Qwen3-1.7B}
\label{sec:act1_qwen}

We first evaluate bidirectional gain installation between pretrained floating-point Qwen3-1.7B ($F(\mathbf{1})$, archive label $\mathrm{FP}_0$) and its asymmetric GPTQ INT3 counterpart ($Q(\mathbf{1})$, archive label $Q_0$; group size $128$, block size $128$, Hessian damping $0.01$, \texttt{act\_order=True} mapped to canonical input-column groups, no MSE clipping) \citep{frantar2023gptq,qwen2025qwen3}. Both donors optimize $393{,}216$ log-scales across four evenly spaced MLP down-projections (layers $6, 13, 20, 27$) for $600$ Adam steps ($\beta=(0.9,0.999)$, $\epsilon=10^{-8}$, zero weight decay, batch order seed $20261012$) on $592$ ARC items, $264$ WikiText-103 blocks, and $260$ C4 blocks with fixed objective weights $(0.0806, 0.4068, 0.5126)$ \citep{kingma2015adam,clark2018arc,hinton2015distilling,merity2016pointer,raffel2020t5}. Crucially, both donors follow the identical two-phase step-size schedule: steps $1$--$200$ use constant $\text{lr}=0.01$ (yielding intermediate checkpoints $F_{200}$ and $Q_{200}$), and steps $201$--$600$ continue from the exact $200$-step state, optimizer moments, and data cursors with linear decay $\text{lr}(t) = 0.01 - 0.009(t - 200)/400$ down to $0.001$ (yielding $600$-step donor gains $a_F$ and $a_Q$; Appendix~\ref{app:protocol}).

\begin{figure}[!t]
\centering
\includegraphics[width=\linewidth]{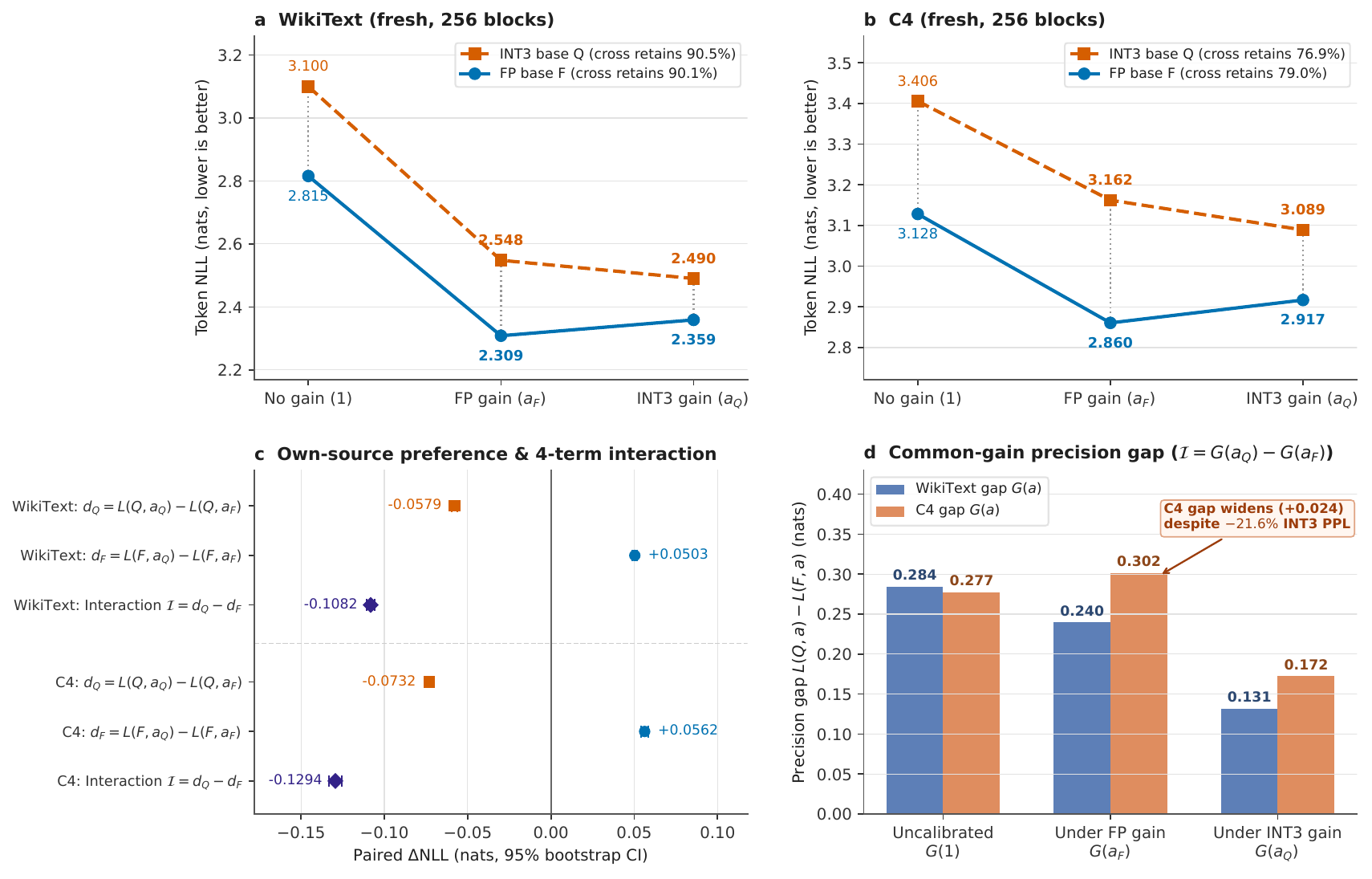}
\caption{\textbf{Bidirectional scale-gain transfer and common-gain precision gaps on fresh unseen Qwen3-1.7B text.} \textbf{a,b}, Token NLL across uncalibrated ($\mathbf{1}$), floating-point donor ($a_F$), and INT3 donor ($a_Q$) gains on fresh WikiText ($256$ blocks, $139$ unseen articles) and C4 ($256$ blocks, $1{,}103$ unseen documents); vertical dotted lines indicate $G(a) = L(Q,a) - L(F,a)$. \textbf{c}, Paired $95\%$ bootstrap intervals for own-source preferences $d_Q, d_F$ and four-term interaction $\mathcal{I} = d_Q - d_F$. \textbf{d}, Common-gain precision gaps: on C4, installing $a_F$ reduces INT3 perplexity by $21.6\%$ ($30.13 \to 23.62$) while widening $G(a_F)$ above $G(\mathbf{1})$ by $+0.024\text{ nats}$ via~\eqref{eq:gap_paradox}.}
\label{fig:crossed}
\end{figure}

Because internal development splits (\textsc{Validation} and \textsc{Replay}) informed earlier analysis, primary text evaluation uses strictly isolated \emph{fresh confirmation cohorts}: $256$ contiguous $2{,}048$-token blocks from $139$ unseen WikiText-103 articles (excluding all $932$ project-consumed articles and official validation/test titles) and $256$ blocks from $1{,}103$ unseen C4 shard-$7$ documents (disjoint from all $319{,}032$ documents in shards $0$--$6$). All six states---$F(\mathbf{1}), F(a_F), F(a_Q), Q(\mathbf{1}), Q(a_F), Q(a_Q)$---are scored in FP32 SDPA arithmetic with zero additional tuning (Table~\ref{tab:qwen_fresh}, Figure~\ref{fig:crossed}):
\begin{enumerate}[leftmargin=1.6em,itemsep=2pt,topsep=2pt]
  \item \textbf{Substantial bidirectional benefit transfer.} Installing the floating-point donor gain $a_F$---trained with zero quantization error---onto INT3 $Q(\mathbf{1})$ reduces fresh WikiText NLL by $-0.5518\text{ nats}$ ($-42.41\%$ perplexity, $22.19 \to 12.78$) and fresh C4 NLL by $-0.2435\text{ nats}$ ($-21.61\%$ perplexity, $30.13 \to 23.62$), achieving retained benefit ratios $R_Q$ of $\mathbf{90.51\%}$ and $\mathbf{76.89\%}$. Conversely, installing the INT3-trained gain $a_Q$ onto clean floating-point $F(\mathbf{1})$ cuts WikiText and C4 perplexities by $36.66\%$ and $19.06\%$, retaining $\mathbf{90.07\%}$ and $\mathbf{78.99\%}$ of floating-point self-calibration ($R_F$). Earlier development splits closely match these fresh confirmation ratios (\textsc{Replay}: $90.10\%/78.03\%$ on $Q$, $90.21\%/80.07\%$ on $F$; Appendix Table~\ref{tab:a_qwen_dev}).
  \item \textbf{Reciprocal own-source preference.} Despite large transferable gains, each recipient significantly prefers the donor trained in its own representation: $d_Q = -0.0579\text{ nats}$ on WikiText and $-0.0732\text{ nats}$ on C4 (favoring $a_Q$ on $Q$), whereas $d_F = +0.0503\text{ nats}$ and $+0.0562\text{ nats}$ (favoring $a_F$ on $F$). The crossed interaction $\mathcal{I} = d_Q - d_F$ is strictly negative on both domains ($-0.1082$ and $-0.1294\text{ nats}$), ruling out a representation-independent ordering of the two donor updates.
  \item \textbf{Divergent common-gain gap versus absolute improvement on C4.} On WikiText, $a_F$ narrows the precision gap from $G(\mathbf{1}) = 0.2844$ to $G(a_F) = 0.2395\text{ nats}$, and $a_Q$ reduces it further to $0.1313\text{ nats}$. On C4, however, installing $a_F$ \emph{increases} $G(a)$ from $0.2774$ to $0.3017\text{ nats}$ ($+0.0242\text{ nats}$, Figure~\ref{fig:crossed}d) even as $Q$'s C4 perplexity drops by $21.6\%$: by~\eqref{eq:gap_paradox}, clean $F$ extracts $-0.2677\text{ nats}$ from $a_F$ while quantized $Q$ extracts $-0.2435\text{ nats}$, before switching to $a_Q$ shifts the gap by $\mathcal{I} = -0.1294\text{ nats}$ to $0.1723\text{ nats}$.
\end{enumerate}

\begin{table}[!t]
\centering
\caption{Bidirectional scale-gain transfer on fresh unseen Qwen3-1.7B cohorts ($256$ blocks of $2{,}048$ tokens per domain; $10{,}000$ paired bootstrap draws). Archive identifiers ($\mathrm{FP}_0, \mathrm{FP}_{\mathrm{SELF}}, \mathrm{FP}_{\mathrm{FROM\_Q}}, Q_0, Q_{\mathrm{FROM\_FP}}, Q_{\mathrm{SELF}}$) and $6$-decimal development-split replications appear in Appendix~\ref{app:qwen_archival} (Tables~\ref{tab:a_qwen_means}--\ref{tab:a_qwen_dev}).}
\label{tab:qwen_fresh}
\setlength{\tabcolsep}{5pt}
\renewcommand{\arraystretch}{1.15}
\small
\fittable{%
\begin{tabular}{@{}llrrrrrr@{}}
\toprule
\multicolumn{2}{@{}l}{Model State / Contrast} & \multicolumn{3}{c}{WikiText (Fresh, $139$ articles)} & \multicolumn{3}{c}{C4 (Fresh, $1{,}103$ docs)} \\
\cmidrule(lr){3-5}\cmidrule(l){6-8}
Recipient & Gain Installed & NLL $\downarrow$ & PPL $\downarrow$ & Retained $R_B$ & NLL $\downarrow$ & PPL $\downarrow$ & Retained $R_B$ \\
\midrule
FP base $F$ & None $F(\mathbf{1})$ [$\mathrm{FP}_0$] & $2.8155$ & $16.70$ & --- & $3.1282$ & $22.83$ & --- \\
FP base $F$ & Own FP gain $F(a_F)$ [$\mathrm{FP}_{\mathrm{SELF}}$] & $2.3085$ & $10.06$ & $100.0\%$ & $2.8605$ & $17.47$ & $100.0\%$ \\
FP base $F$ & Cross INT3 gain $F(a_Q)$ [$\mathrm{FP}_{\mathrm{FROM\_Q}}$] & $2.3589$ & $10.58$ & $\mathbf{90.07\%}$ & $2.9167$ & $18.48$ & $\mathbf{78.99\%}$ \\
\addlinespace[2pt]
INT3 base $Q$ & None $Q(\mathbf{1})$ [$Q_0$] & $3.0998$ & $22.19$ & --- & $3.4056$ & $30.13$ & --- \\
INT3 base $Q$ & Cross FP gain $Q(a_F)$ [$Q_{\mathrm{FROM\_FP}}$] & $2.5480$ & $12.78$ & $\mathbf{90.51\%}$ & $3.1622$ & $23.62$ & $\mathbf{76.89\%}$ \\
INT3 base $Q$ & Own INT3 gain $Q(a_Q)$ [$Q_{\mathrm{SELF}}$] & $2.4902$ & $12.06$ & $100.0\%$ & $3.0890$ & $21.96$ & $100.0\%$ \\
\midrule
\multicolumn{2}{@{}l}{\emph{Paired Contrasts \& Precision Gaps (nats)}} & \multicolumn{3}{c}{Estimate $[95\%\text{ Bootstrap CI}]$} & \multicolumn{3}{c}{Estimate $[95\%\text{ Bootstrap CI}]$} \\
\midrule
\multicolumn{2}{@{}l}{Transfer gain on $Q$: $L(Q,a_F) - L(Q,\mathbf{1})$} & \multicolumn{3}{c}{$-0.5518\;[-0.5649,\,-0.5387]$} & \multicolumn{3}{c}{$-0.2435\;[-0.2504,\,-0.2362]$} \\
\multicolumn{2}{@{}l}{Transfer gain on $F$: $L(F,a_Q) - L(F,\mathbf{1})$} & \multicolumn{3}{c}{$-0.4566\;[-0.4655,\,-0.4476]$} & \multicolumn{3}{c}{$-0.2114\;[-0.2176,\,-0.2053]$} \\
\multicolumn{2}{@{}l}{$Q$ source preference: $d_Q = L(Q,a_Q) - L(Q,a_F)$} & \multicolumn{3}{c}{$-0.0579\;[-0.0596,\,-0.0560]$} & \multicolumn{3}{c}{$-0.0732\;[-0.0753,\,-0.0709]$} \\
\multicolumn{2}{@{}l}{$F$ source preference: $d_F = L(F,a_Q) - L(F,a_F)$} & \multicolumn{3}{c}{$+0.0503\;[+0.0489,\,+0.0518]$} & \multicolumn{3}{c}{$+0.0562\;[+0.0540,\,+0.0586]$} \\
\multicolumn{2}{@{}l}{Interaction: $\mathcal{I} = d_Q - d_F = G(a_Q)-G(a_F)$} & \multicolumn{3}{c}{$-0.1082\;[-0.1105,\,-0.1058]$} & \multicolumn{3}{c}{$-0.1294\;[-0.1333,\,-0.1256]$} \\
\multicolumn{2}{@{}l}{Precision gaps: $G(\mathbf{1}) \;\to\; G(a_F) \;\to\; G(a_Q)$} & \multicolumn{3}{c}{$0.2844 \;\to\; 0.2395 \;\to\; 0.1313$} & \multicolumn{3}{c}{$0.2774 \;\to\; \mathbf{0.3017} \;\to\; 0.1723$} \\
\bottomrule
\end{tabular}}
\end{table}

\subsection{Coordinate Alignment, Layer Placement, and Non-Local Context}
\label{sec:act1_controls}

\begin{figure}[!t]
\centering
\includegraphics[width=\linewidth]{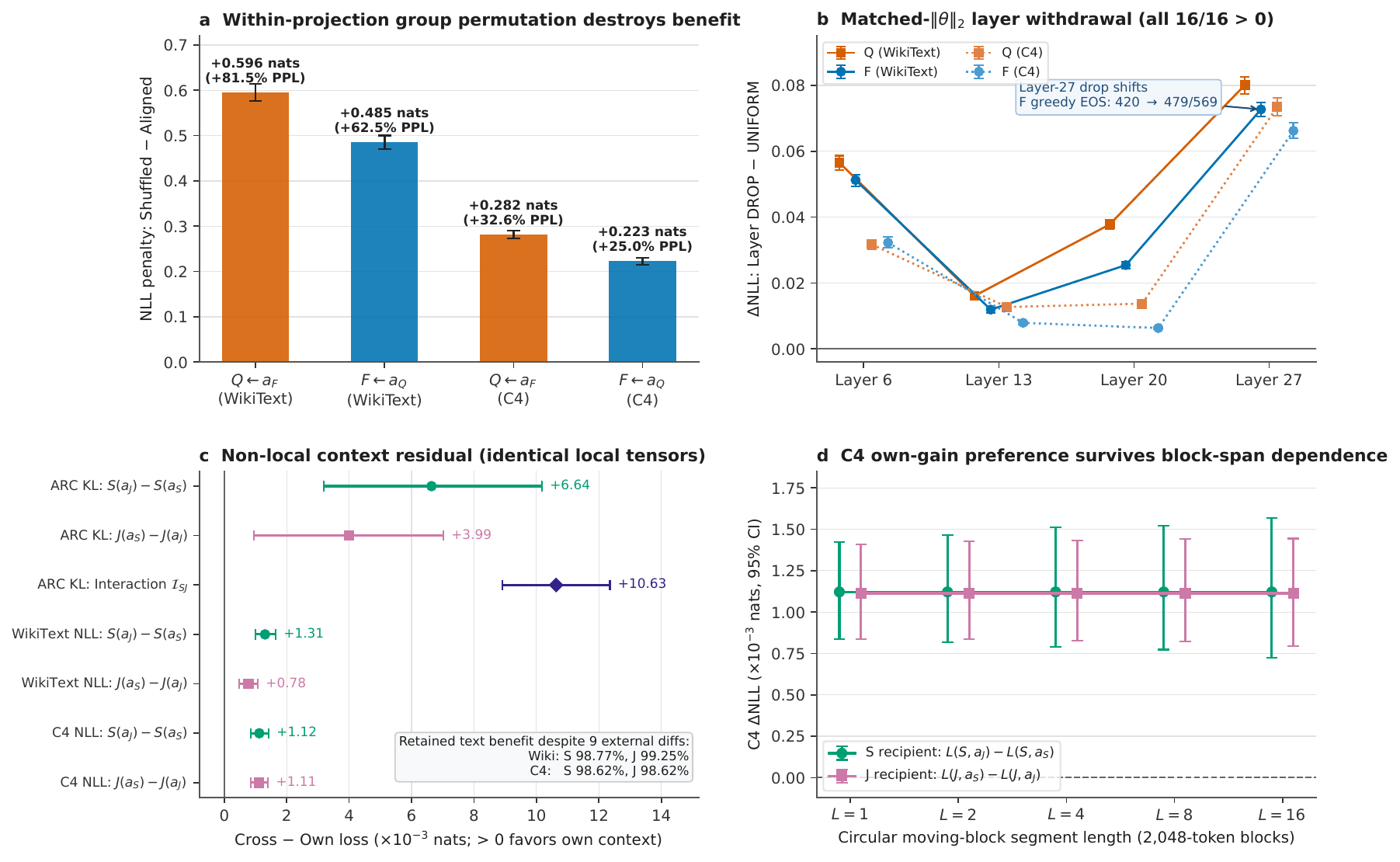}
\caption{\textbf{Coordinate alignment, depth sensitivity, and non-local context controls.} \textbf{a}, Permuting group gains within each calibrated Qwen3-1.7B projection preserves exact per-layer gain distributions yet destroys cross-transfer ($+25.0\%$ to $+81.5\%$ PPL penalty). \textbf{b}, Withdrawing any single calibrated layer ($6, 13, 20, 27$) on $200$-step checkpoints incurs strictly higher NLL than distributing the exact same log-gain $\ell_2$ shrinkage across all four layers ($\text{DROP} - \text{UNIFORM} > 0$ in all $16/16$ cells) and shifts layer-$27$ greedy EOS counts. \textbf{c}, On Llama-3.2-3B INT3 recipients $S$ and $J$ with \emph{bit-identical} initial tensors at all four calibrated projections and $9$ external quantization differences elsewhere, swapping gains retains $98.62\%$--$99.25\%$ of text benefit while leaving resolved positive own-context preferences. \textbf{d}, Circular moving-block bootstrap CIs on C4 ($L \in \{1,2,4,8,16\}$) confirm robustness to multi-block document spans.}
\label{fig:controls}
\end{figure}

To test what structural and contextual properties govern scale transfer and recipient preference, we run three targeted controls (Figure~\ref{fig:controls}).

\paragraph{Within-projection group permutation.}
To test whether transferred benefit merely reflects coarse per-layer norm or temperature rescaling, we randomly permute the $98{,}304$ group multipliers $(o, c)$ independently within each of the four calibrated Qwen3-1.7B projections, preserving every layer's exact gain multiset, mean, and $\|\theta_l\|_2$ while breaking group-to-channel correspondence. Group permutation eliminates nearly all transferred benefit (Table~\ref{tab:qwen_controls}, Figure~\ref{fig:controls}a): relative to aligned installation, shuffled gains degrade $Q(a_F)$ by $+0.5958\text{ nats}$ ($+81.46\%$ PPL) on WikiText and $+0.2821\text{ nats}$ ($+32.60\%$ PPL) on C4 (exceeding uncalibrated $Q_0$'s $3.2860$ \textsc{Replay} C4 NLL at $3.3227$), and degrade $F(a_Q)$ by $+0.4854$ and $+0.2231\text{ nats}$. Cross-precision transfer therefore requires exact $(o, c)$ coordinate alignment rather than layerwise summary statistics.

\begin{table}[!t]
\centering
\caption{Coordinate-alignment and depth-placement controls on Qwen3-1.7B: within-projection group permutation ($600$-step cross-installed gains on \textsc{Replay}) and norm-matched single-layer withdrawal ($200$-step checkpoints on \textsc{Validation}; $\alpha_l = \sqrt{1 - \|\theta_l\|_2^2/\|\theta\|_2^2}$). Full $16$-cell $6$-decimal placement ledger appears in Appendix~\ref{app:qwen_archival} (Table~\ref{tab:a_placement}).}
\label{tab:qwen_controls}
\setlength{\tabcolsep}{5pt}
\renewcommand{\arraystretch}{1.15}
\small
\fittable{%
\begin{tabular}{@{}llrrr@{}}
\toprule
Control Experiment & Condition / Layer & WikiText $\Delta$NLL $[95\%\text{ CI}]$ & C4 $\Delta$NLL $[95\%\text{ CI}]$ & PPL Shift \\
\midrule
\multirow{2}{*}{Group Permutation} & $Q(a_F)$: $\text{Shuffled} - \text{Aligned}$ & $+0.5958\;[+0.5767,\,+0.6148]$ & $+0.2821\;[+0.2737,\,+0.2904]$ & $+81.5\% / +32.6\%$ \\
 & $F(a_Q)$: $\text{Shuffled} - \text{Aligned}$ & $+0.4854\;[+0.4706,\,+0.5002]$ & $+0.2231\;[+0.2153,\,+0.2305]$ & $+62.5\% / +25.0\%$ \\
\midrule
\multirow{5}{*}{\shortstack[l]{Matched-$\|\theta\|_2$ Layer\\Withdrawal\\($\text{DROP} - \text{UNIFORM}$)}} & $Q_{200}$: Layer $13$ (min both) & $+0.0161\;[+0.0152,\,+0.0171]$ & $+0.0127\;[+0.0121,\,+0.0134]$ & $+1.62\% / +1.28\%$ \\
 & $Q_{200}$: Layer $27$ (max both) & $+0.0801\;[+0.0773,\,+0.0827]$ & $+0.0734\;[+0.0707,\,+0.0761]$ & $+8.34\% / +7.62\%$ \\
 & $F_{200}$: Layer $13$ (min Wiki) & $+0.0119\;[+0.0109,\,+0.0128]$ & $+0.0079\;[+0.0075,\,+0.0084]$ & $+1.20\% / +0.79\%$ \\
 & $F_{200}$: Layer $20$ (min C4) & $+0.0254\;[+0.0244,\,+0.0264]$ & $+0.0063\;[+0.0059,\,+0.0067]$ & $+2.58\% / +0.63\%$ \\
 & $F_{200}$: Layer $27$ (max both)$^\dagger$ & $+0.0727\;[+0.0705,\,+0.0748]$ & $+0.0662\;[+0.0639,\,+0.0686]$ & $+7.54\% / +6.85\%$ \\
\bottomrule
\end{tabular}}
\par\vspace{2pt}\raggedright{\footnotesize $^\dagger$Withdrawing Layer $27$ on $F_{200}$ shifts $64$-token greedy ARC EOS terminations from $420/569$ (\textsc{Uniform}) to $479/569$ (\textsc{Drop}).\par}
\end{table}

\paragraph{Norm-matched layer withdrawal.}
To test whether sensitivity is uniform across depth at fixed log-gain norm, we take $Q_{200}$ and $F_{200}$ and compare \textsc{Drop}-$l$ ($\theta_l \leftarrow 0$, retaining the other three layers) against \textsc{Uniform}-$l$, which scales all four layers by $\alpha_l = \sqrt{1 - \|\theta_l\|_2^2/\|\theta\|_2^2}$ so $\|\theta_{\mathrm{DROP},l}\|_2 = \|\theta_{\mathrm{UNIFORM},l}\|_2$. Across both bases, all four layers, and both domains ($16/16$ cells), $\text{DROP} - \text{UNIFORM}$ is strictly positive (Table~\ref{tab:qwen_controls}, Figure~\ref{fig:controls}b), peaking at layer $27$ ($+0.0801\text{ nats}$ on $Q_{200}$, $+0.0727\text{ nats}$ on $F_{200}$ on WikiText) and layer $6$ ($+0.0566$ and $+0.0512\text{ nats}$). Concentrating norm removal at layer $27$ on $F_{200}$ also increases early EOS terminations in $64$-token greedy ARC generations from $420/569$ to $479/569$, showing that layer placement materially affects both token loss and generation stopping at matched parameter norm.

\begin{table}[!t]
\centering
\caption{Isolating non-local network context on Llama-3.2-3B INT3 ($4$ calibrated \texttt{mlp.down\_proj} sites at layers $6, 13, 20, 27$; $786{,}432$ scales). Configurations $S$ and $J$ have \emph{bit-identical} initial codes, scales, and zero points at all four calibrated sites ($\delta_g^{(S,J)} \equiv 0$) and differ only at $9$ uncalibrated projections elsewhere (Appendix~\ref{app:llama_context}, Table~\ref{tab:a_llama_sites}).}
\label{tab:llama_context}
\setlength{\tabcolsep}{5pt}
\renewcommand{\arraystretch}{1.15}
\small
\fittable{%
\begin{tabular}{@{}lrrrrr@{}}
\toprule
State / Paired Contrast ($10{,}000$ draws) & ARC KL $\downarrow$ & WikiText NLL $\downarrow$ & C4 NLL $\downarrow$ & Wiki $R_B$ & C4 $R_B$ \\
\midrule
$S(\mathbf{1})$ uncalibrated [$S_0$] & $0.1774$ & $2.3697$ & $2.7866$ & --- & --- \\
$S(a_S)$ own gain [$S_{\mathrm{SELF}}$] & $0.1000$ & $2.2634$ & $2.7056$ & $100.0\%$ & $100.0\%$ \\
$S(a_J)$ cross gain [$S_{\mathrm{FROM\_J}}$] & $0.1067$ & $2.2647$ & $2.7067$ & $\mathbf{98.77\%}$ & $\mathbf{98.62\%}$ \\
\addlinespace[2pt]
$J(\mathbf{1})$ uncalibrated [$J_0$] & $0.1731$ & $2.3694$ & $2.7853$ & --- & --- \\
$J(a_J)$ own gain [$J_{\mathrm{SELF}}$] & $0.1060$ & $2.2646$ & $2.7048$ & $100.0\%$ & $100.0\%$ \\
$J(a_S)$ cross gain [$J_{\mathrm{FROM\_S}}$] & $0.1099$ & $2.2654$ & $2.7059$ & $\mathbf{99.25\%}$ & $\mathbf{98.62\%}$ \\
\midrule
\emph{Cross $-$ Own ($\times 10^{-3}\text{ nats}$)} & \multicolumn{2}{c}{ARC $\Delta$KL $[95\%\text{ CI}]$} & \multicolumn{2}{c}{WikiText $\Delta$NLL $[95\%\text{ CI}]$} & C4 $\Delta$NLL $[95\%\text{ CI}]$ \\
\midrule
$S$ recipient: $L(S,a_J) - L(S,a_S)$ & \multicolumn{2}{c}{$+6.64\;[+3.19,\,+10.18]$} & \multicolumn{2}{c}{$+1.31\;[+0.99,\,+1.65]$} & $+1.12\;[+0.85,\,+1.42]$ \\
$J$ recipient: $L(J,a_S) - L(J,a_J)$ & \multicolumn{2}{c}{$+3.99\;[+0.96,\,+7.02]$} & \multicolumn{2}{c}{$+0.78\;[+0.49,\,+1.07]$} & $+1.11\;[+0.85,\,+1.40]$ \\
Context interaction $\mathcal{I}_{SJ}$ & \multicolumn{2}{c}{$+10.63\;[+8.92,\,+12.36]$} & \multicolumn{2}{c}{$+2.09\;[+1.98,\,+2.21]$} & $+2.23\;[+1.93,\,+2.65]$ \\
\bottomrule
\end{tabular}}
\end{table}

\paragraph{Isolating non-local context with bit-identical local tensors.}
In Qwen's $F \leftrightarrow Q$ experiment, local initial weight differences ($\delta_g \neq 0$) coexist with surrounding-layer quantization differences. To test whether surrounding uncalibrated layers alone induce recipient-specific gain preferences, we evaluate two fixed INT3 configurations on Llama-3.2-3B \citep{meta2024llama32} ($28$ layers, $196$ group-$128$ projections): $S$ (\textsc{Static31}) and $J$ (\textsc{Joint39}). At all four calibrated \texttt{mlp.down\_proj} projections (layers $6, 13, 20, 27$; $786{,}432$ scales), $S$ and $J$ share \emph{bit-for-bit identical} initial integer codes $q_g$, zero points $z_g$, and scales $s_{g,0}$---so their local initial difference $\delta_g^{(S,J)} := w_{S,g,0} - w_{J,g,0} \equiv 0$---while differing at exactly nine uncalibrated projections in layers $4, 7, 8, 12, 14, 15, 17$ (switches among GPTQ $1\%$, GPTQ $0.1\%$, and round-to-nearest; Appendix Table~\ref{tab:a_llama_sites}). Both donors run $200$ Adam steps at constant $\text{lr}=0.01$ (seed $20261002$, weights $(0.0266, 0.3424, 0.6310)$) on $592$ ARC items, $28$ WikiText blocks, and $256$ C4 blocks.

Swapping $a_S$ and $a_J$ retains $98.77\%$ ($S$) and $99.25\%$ ($J$) of WikiText NLL benefit and $98.62\%$ on C4, yet every pointwise $95\%$ bootstrap CI for $\text{Cross} - \text{Own}$ loss on ARC KL ($+6.64 \times 10^{-3}$ on $S$, $+3.99 \times 10^{-3}$ on $J$), WikiText NLL ($+1.31 \times 10^{-3}$, $+0.78 \times 10^{-3}$), and C4 NLL ($+1.12 \times 10^{-3}$, $+1.11 \times 10^{-3}$), plus their interaction $\mathcal{I}_{SJ}$, lies strictly above zero (Table~\ref{tab:llama_context}, Figure~\ref{fig:controls}c).\footnote{Under conservative familywise multiplicity adjustments across all six individual contrasts simultaneously, the widest individual relative interval ($J$'s ARC KL contrast, $+3.99 \times 10^{-3}$) approaches zero while all four text-NLL preferences and all three interaction terms remain strictly positive.} Circular moving-block C4 bootstraps across segment lengths $L \in \{1, 2, 4, 8, 16\}$ confirm that multi-block document spans ($221/1{,}128$ evaluation documents cross block boundaries) do not explain this effect ($[+0.72, +1.57]\times 10^{-3}$ on $S$ and $[+0.79, +1.44]\times 10^{-3}$ on $J$ at $L=16$; Figure~\ref{fig:controls}d, Table~\ref{tab:a_c4_block}). Note that because Qwen ($F \leftrightarrow Q$) and Llama ($S \leftrightarrow J$) differ in architecture, training schedule, pool size, and perturbation magnitude, comparing their retention ranges ($77\%$--$91\%$ vs.\ $\sim 99\%$) does not partition local versus non-local variance within a single model; rather, the $S \leftrightarrow J$ control independently establishes that non-local quantization context alters optimal local scale updates even when $\delta_g^{(S,J)} \equiv 0$.

\section{Full-Scale Calibration Capacity and Step-Size Control}
\label{sec:act2}

\begin{figure}[!t]
\centering
\includegraphics[width=\linewidth]{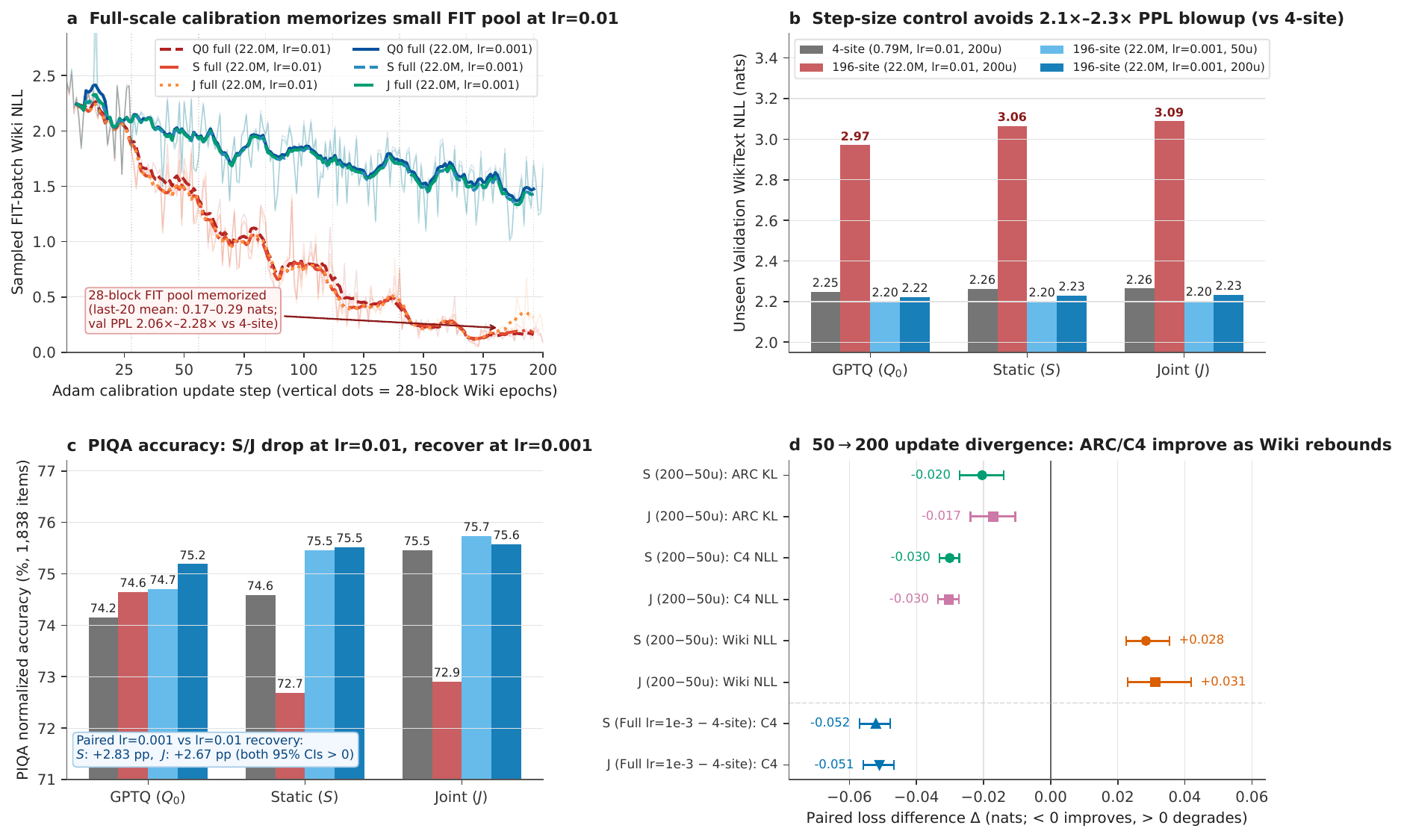}
\caption{\textbf{Full-scale calibration memorization, step-size control, and multi-objective trajectory divergence on Llama-3.2-3B ($22.02\text{M}$ scales).} \textbf{a}, Sampled training-batch WikiText NLL across $200$ Adam steps ($9$-step moving averages; dotted vertical lines mark $28$-block WikiText epochs). At $\text{lr}=0.01$, $22.02\text{M}$ scales memorize the $28$-block pool ($0.17$--$0.29\text{ nats}$). \textbf{b}, Unseen validation WikiText NLL: $196$-site $\text{lr}=0.01$ inflates perplexity $2.06\times$--$2.28\times$ relative to $4$-site calibration, whereas $\text{lr}=0.001$ avoids memorization. \textbf{c}, Downstream PIQA normalized accuracy ($1{,}838$ items) drops by $1.90$--$2.56\text{ pp}$ on $S/J$ at $\text{lr}=0.01$ ($+0.49\text{ pp}$ on $Q_0$) and recovers by $+2.83$ and $+2.67\text{ pp}$ at $\text{lr}=0.001$. \textbf{d}, Advancing from $50$ to $200$ steps at $\text{lr}=0.001$ improves ARC KL and C4 NLL while unseen WikiText NLL rebounds ($+0.028$ to $+0.031\text{ nats}$).}
\label{fig:capacity}
\end{figure}

Because native scale absorption adds zero inference overhead regardless of how many projections are calibrated, a natural extension of $4$-site calibration ($0.79\text{M}$ scales on Llama-3.2-3B) is to unlock all $196$ linear projections ($22{,}020{,}096$ scales, $28\times$ capacity) on the same compact PTQ calibration pool ($592$ ARC items, $28$ WikiText blocks, $256$ C4 blocks). Evaluating $12$ Llama-3.2-3B states across bases $\{Q_0, S, J\}$, site budgets $\{4\text{-site}, 196\text{-site}\}$, learning rates $\{0.01, 0.001\}$, and step counts $\{50, 200\}$ exposes severe small-pool memorization at $\text{lr}=0.01$ and full recovery under step-size control at $\text{lr}=0.001$ (Table~\ref{tab:llama_capacity}, Figure~\ref{fig:capacity}).

\subsection{The Consumed-FIT Pilot Trap and Small-Pool Memorization}
\label{sec:act2_collapse}

When screening learning rates $\{0.001, 0.003, 0.01\}$ on $Q_0$ using a $50$-step pilot without consulting held-out validation data, consumed calibration (\textsc{Fit}) loss ranks $\text{lr}=0.01$ best ($2.1031$, versus $2.1380$ at $\text{lr}=0.003$ and $2.3732$ at $\text{lr}=0.001$; Appendix~\ref{app:capacity_archival}). Most of this advantage comes from lower consumed \textsc{Fit} WikiText NLL ($1.2839$ vs.\ $1.9482\text{ nats}$ at $\text{lr}=0.001$), with a smaller C4 improvement ($2.6315$ vs.\ $2.7008$); together they outweigh worse \textsc{Fit} ARC KL ($0.1121$ vs.\ $0.0722$).

Continuing $196$-site $\text{lr}=0.01$ to $200$ steps---over seven passes through the $28$-block WikiText pool ($57{,}344$ tokens, $<2.6\times 10^{-3}$ tokens per trainable scale)---causes severe overfitting (Figure~\ref{fig:capacity}a--c):
\begin{itemize}[leftmargin=1.6em,itemsep=2pt,topsep=2pt]
  \item \textbf{Training-pool memorization vs.\ validation blowup.} Across steps $181$--$200$, sampled training-batch WikiText NLL collapses from $2.20$--$2.21$ to $0.166\text{ nats}$ ($Q_0$), $0.182\text{ nats}$ ($S$), and $0.291\text{ nats}$ ($J$), while unseen validation WikiText NLL rises to $2.9708$, $3.0630$, and $3.0882\text{ nats}$. Relative to $4$-site $\text{lr}=0.01$ calibration ($2.2487$, $2.2634$, $2.2646\text{ nats}$), these $+0.7221$ to $+0.8236\text{ nats}$ increases correspond to $\mathbf{2.059\times}$, $\mathbf{2.225\times}$, and $\mathbf{2.279\times}$ \textbf{validation perplexity ratios} ($19.51$--$21.94$ vs.\ $9.48$--$9.63$; Table~\ref{tab:llama_capacity}). Unseen C4 NLL also worsens by $+0.1163$ to $+0.1647\text{ nats}$, and ARC KL rises by $+0.0254$ to $+0.1088\text{ nats}$.
  \item \textbf{Base-dependent downstream PIQA degradation.} On held-out PIQA \citep{bisk2019piqa} ($1{,}838$ items), $196$-site $\text{lr}=0.01$ relative to $4$-site calibration leaves uniform GPTQ $Q_0$ statistically unchanged ($+0.49\text{ pp}$ $[-0.76, +1.80]$, $p=0.47$), but significantly degrades mixed-action slates $S$ by $\mathbf{-1.90\text{ pp}}$ ($[-3.32, -0.44]$, $-35$ net items, exact McNemar $p = 0.0113$) and $J$ by $\mathbf{-2.56\text{ pp}}$ ($[-4.03, -1.09]$, $-47$ net items, $p = 9.39\times 10^{-4}$; interaction $S - Q_0$ of $-2.39\text{ pp}$ and $J - Q_0$ of $-3.05\text{ pp}$ in Table~\ref{tab:a_capacity}). Even with frozen integer codes, $22\text{M}$ unconstrained log-scales readily overfit small PTQ text pools.
\end{itemize}

\begin{table}[!t]
\centering
\caption{Capacity, learning-rate, and step-count regimes on Llama-3.2-3B INT3 across bases $Q_0$ (uniform GPTQ), $S$ (\textsc{Static31}), and $J$ (\textsc{Joint39}). ``$4$-site'' calibrates $0.79\text{M}$ scales; ``$196$-site'' calibrates all $22.02\text{M}$ scales on the identical calibration pool. Full $6$-decimal contrasts and four-term interactions appear in Appendix~\ref{app:capacity_archival} (Table~\ref{tab:a_capacity}).}
\label{tab:llama_capacity}
\setlength{\tabcolsep}{4.8pt}
\renewcommand{\arraystretch}{1.15}
\small
\fittable{%
\begin{tabular}{@{}llrrrrrr@{}}
\toprule
Regime (Sites, LR, Updates) & Base & ARC KL $\downarrow$ & Wiki NLL $\downarrow$ & Wiki PPL $\downarrow$ & C4 NLL $\downarrow$ & PIQA Norm $\uparrow$ & PIQA Raw $\uparrow$ \\
\midrule
\multirow{3}{*}{$4$-site ($0.79\text{M}$, $\text{lr}=10^{-2}$, $200\text{u}$)} & $Q_0$ & $0.1030$ & $2.2487$ & $9.48$ & $2.7061$ & $74.16\%$ & $73.56\%$ \\
 & $S$ & $0.1000$ & $2.2634$ & $9.62$ & $2.7056$ & $74.59\%$ & $74.48\%$ \\
 & $J$ & $0.1060$ & $2.2646$ & $9.63$ & $2.7048$ & $75.46\%$ & $74.59\%$ \\
\addlinespace[2pt]
\multirow{3}{*}{$196$-site ($22.0\text{M}$, $\text{lr}=10^{-2}$, $200\text{u}$)} & $Q_0$ & $0.1283$ & $2.9708$ & $19.51$ & $2.8290$ & $74.65\%$ & $74.59\%$ \\
 & $S$ & $0.2089$ & $3.0630$ & $21.39$ & $2.8219$ & $72.69\%$ & $72.74\%$ \\
 & $J$ & $0.1894$ & $3.0882$ & $21.94$ & $2.8695$ & $72.91\%$ & $72.47\%$ \\
\addlinespace[2pt]
\multirow{3}{*}{$196$-site ($22.0\text{M}$, $\text{lr}=10^{-3}$, $50\text{u}$)} & $Q_0$ & $0.0933$ & $2.2017$ & $9.04$ & $2.6868$ & $74.70\%$ & $73.83\%$ \\
 & $S$ & $0.0966$ & $2.2013$ & $9.04$ & $2.6836$ & $75.46\%$ & $75.19\%$ \\
 & $J$ & $0.0964$ & $2.2034$ & $9.06$ & $2.6841$ & $75.73\%$ & $75.24\%$ \\
\addlinespace[2pt]
\multirow{3}{*}{$196$-site ($22.0\text{M}$, $\text{lr}=10^{-3}$, $200\text{u}$)} & $Q_0$ & $0.0810$ & $2.2214$ & $9.22$ & $2.6534$ & $75.19\%$ & $74.92\%$ \\
 & $S$ & $0.0762$ & $2.2297$ & $9.30$ & $2.6535$ & $75.52\%$ & $75.24\%$ \\
 & $J$ & $0.0792$ & $2.2346$ & $9.34$ & $2.6537$ & $75.57\%$ & $75.73\%$ \\
\midrule
\multicolumn{2}{@{}l}{\emph{Key Paired Regime Contrasts ($95\%$ CI)}} & \multicolumn{2}{c}{ARC $\Delta$KL} & \multicolumn{2}{c}{WikiText $\Delta$NLL} & C4 $\Delta$NLL & PIQA $\Delta$Norm \\
\midrule
\multicolumn{2}{@{}l}{Overfit: $Q_0$ ($196\text{-site }10^{-2} - 4\text{-site}$)} & \multicolumn{2}{c}{$+0.0254\;[+0.010,\,+0.040]$} & \multicolumn{2}{c}{$+0.7221\;[+0.678,\,+0.760]$} & $+0.1230$ & $+0.49\text{ pp}$ \\
\multicolumn{2}{@{}l}{Overfit: $S$ ($196\text{-site }10^{-2} - 4\text{-site}$)} & \multicolumn{2}{c}{$+0.1088\;[+0.090,\,+0.129]$} & \multicolumn{2}{c}{$+0.7996\;[+0.741,\,+0.841]$} & $+0.1163$ & $-1.90\text{ pp}^*$ \\
\multicolumn{2}{@{}l}{Overfit: $J$ ($196\text{-site }10^{-2} - 4\text{-site}$)} & \multicolumn{2}{c}{$+0.0835\;[+0.065,\,+0.102]$} & \multicolumn{2}{c}{$+0.8236\;[+0.762,\,+0.868]$} & $+0.1647$ & $-2.56\text{ pp}^{**}$ \\
\multicolumn{2}{@{}l}{Recovery: $S$ ($196\text{-site }10^{-3} - 10^{-2}$, $200\text{u}$)} & \multicolumn{2}{c}{$-0.1327\;[-0.152,\,-0.115]$} & \multicolumn{2}{c}{$-0.8332\;[-0.853,\,-0.812]$} & $-0.1685$ & $+2.83\text{ pp}^{**}$ \\
\multicolumn{2}{@{}l}{Recovery: $J$ ($196\text{-site }10^{-3} - 10^{-2}$, $200\text{u}$)} & \multicolumn{2}{c}{$-0.1102\;[-0.128,\,-0.093]$} & \multicolumn{2}{c}{$-0.8536\;[-0.878,\,-0.828]$} & $-0.2158$ & $+2.67\text{ pp}^{**}$ \\
\multicolumn{2}{@{}l}{Operating point: $S$ ($196\text{-site }10^{-3}\text{ }200\text{u} - 4\text{-site}$)} & \multicolumn{2}{c}{$-0.0238\;[-0.031,\,-0.017]$} & \multicolumn{2}{c}{$-0.0336\;[-0.085,\,-0.001]$} & $-0.0521$ & $+0.92\text{ pp}^*$ \\
\multicolumn{2}{@{}l}{Trajectory: $S$ ($200\text{u} - 50\text{u}$ at $\text{lr}=10^{-3}$)} & \multicolumn{2}{c}{$-0.0204\;[-0.027,\,-0.014]$} & \multicolumn{2}{c}{$+0.0284\;[+0.023,\,+0.035]$} & $-0.0301$ & $+0.05\text{ pp}$ \\
\multicolumn{2}{@{}l}{Trajectory: $J$ ($200\text{u} - 50\text{u}$ at $\text{lr}=10^{-3}$)} & \multicolumn{2}{c}{$-0.0172\;[-0.024,\,-0.011]$} & \multicolumn{2}{c}{$+0.0312\;[+0.023,\,+0.042]$} & $-0.0304$ & $-0.16\text{ pp}$ \\
\bottomrule
\end{tabular}}
\par\vspace{2pt}\raggedright{\footnotesize Exact two-sided McNemar test on $1{,}838$ paired PIQA items: $^*p < 0.05$, $^{**}p < 10^{-3}$ (full intervals in Table~\ref{tab:a_capacity}).\par}
\end{table}

\subsection{Step-Size Control and Asynchronous Multi-Objective Trajectories}
\label{sec:act2_recovery}

Lowering the log-scale Adam learning rate to $\text{lr}=0.001$ keeps step-$200$ training-batch WikiText NLL at $1.42$--$1.46\text{ nats}$ (Figure~\ref{fig:capacity}a) and restores held-out performance:
\begin{enumerate}[leftmargin=1.6em,itemsep=2pt,topsep=2pt]
  \item \textbf{Within-capacity recovery and comparison to $4$-site calibration.} At fixed $196$-site capacity and $200$ updates, $\text{lr}=0.001$ versus $\text{lr}=0.01$ reduces validation WikiText NLL by $-0.8332\text{ nats}$ ($S$) and $-0.8536\text{ nats}$ ($J$), C4 NLL by $-0.1685$ and $-0.2158\text{ nats}$, and ARC KL by $-0.1327$ and $-0.1102\text{ nats}$, while recovering $\mathbf{+2.83\text{ pp}}$ ($[+1.41, +4.24]$, $p = 1.19 \times 10^{-4}$) on $S$ and $\mathbf{+2.67\text{ pp}}$ ($[+1.20, +4.13]$, $p = 4.50 \times 10^{-4}$) on $J$ in normalized PIQA accuracy. Relative to the $4$-site $\text{lr}=0.01$ operating point, $196$-site $\text{lr}=0.001$ ($200\text{u}$) consistently reduces ARC KL ($-0.0220$ to $-0.0268$) and C4 NLL ($-0.0510$ to $-0.0526$) across all three bases, with smaller pointwise PIQA shifts ($+0.92\text{ pp}$, $p=0.043$ normalized on $S$; $+0.11\text{ pp}$ normalized and $+1.14\text{ pp}$, $p=0.0099$ raw on $J$).\footnote{Between $196$-site $\text{lr}=0.001$ and $4$-site $\text{lr}=0.01$, both site count and step size change; the single-factor step-size effect at $196$ sites is isolated by $\text{lr}=0.001$ vs.\ $0.01$, and the single-factor capacity effect at $\text{lr}=0.01$ is isolated by $196$-site vs.\ $4$-site.}
  \item \textbf{Asynchronous $50 \to 200$ step trajectories.} Even at $\text{lr}=0.001$, the three validation losses peak at different update horizons (Figure~\ref{fig:capacity}d): between $50$ and $200$ steps, unseen ARC KL ($-0.0204$ on $S$, $-0.0172$ on $J$) and C4 NLL ($-0.0301$, $-0.0304$) continue improving, whereas unseen WikiText NLL rebounds upward by $+0.0284\text{ nats}$ ($[+0.0225, +0.0354]$) on $S$ and $+0.0312\text{ nats}$ ($[+0.0230, +0.0419]$) on $J$. While this ordering aligns qualitatively with the much smaller $28$-block WikiText pool ($>7.1$ passes at $200$ steps vs.\ $0.78$ C4 passes), pool size and domain characteristics co-vary across components, so per-component validation tracking is necessary in practice.
  \item \textbf{Convergence across initial quantization slates.} At $50$ updates ($\text{lr}=0.001$), $S$ and $J$ lead $Q_0$ on raw PIQA ($+1.36\text{ pp}$, $p=0.014$; $+1.41\text{ pp}$, $p=0.016$) and C4 NLL ($-0.0033$, $-0.0028$). By $200$ updates, $Q_0, S, J$ converge to $2.6534$--$2.6537$ C4 NLL and $75.19\%$--$75.57\%$ normalized PIQA ($J - S = +0.05\text{ pp}$ $[-0.71, +0.82]$), indicating that step-size-controlled full-network scale updates narrow initial differences among discrete PTQ slates.
\end{enumerate}

\section{Decision Decoupling, Readout Sign Reversal, and Bias Subtraction}
\label{sec:act3}

Sections~\ref{sec:act1}--\ref{sec:act2} evaluate continuous proxy losses alongside PIQA. Because deployed models are queried through discrete scoring rules and chat templates that differ from plain-text calibration objectives, we next audit how identical scale-calibrated checkpoints behave across PIQA normalization conventions and across a $24$-endpoint ARC interface matrix.

\subsection{Length-Normalization Sensitivity on Downstream PIQA}
\label{sec:act3_piqa}

Standard evaluation harnesses report both character-length-normalized accuracy (\texttt{acc\_norm}, $\arg\max_c |y_c|_{\mathrm{char}}^{-1} \log p_\theta(y_c \mid x)$) and raw log-likelihood accuracy (\texttt{acc}, $\arg\max_c \log p_\theta(y_c \mid x)$) on PIQA's $1{,}838$ validation items \citep{bisk2019piqa}. Across Qwen3-1.7B and Llama-3.2-3B $4$-site states, discrete decision gains decouple from text NLL (Table~\ref{tab:piqa_decoupling}):
\begin{itemize}[leftmargin=1.6em,itemsep=2pt,topsep=2pt]
  \item On Qwen INT3 $Q$, cross-installing $a_F$ cuts fresh WikiText and C4 perplexities by $42.4\%$ and $21.6\%$ (Table~\ref{tab:qwen_fresh}), yet leaves raw PIQA accuracy flat ($67.79\% \to 67.63\%$, $-0.16\text{ pp}$ $[-1.20, +0.87]$, $p=0.84$) and normalized accuracy unresolved ($+1.03\text{ pp}$ $[-0.16, +2.29]$, $p=0.12$). By contrast, own-source calibration $Q(a_Q)$ significantly improves both \texttt{acc\_norm} ($+1.90\text{ pp}$ $[+0.60, +3.21]$, $p=0.0055$) and \texttt{acc} ($+1.69\text{ pp}$ $[+0.44, +2.94]$, $p=0.011$), outperforming $Q(a_F)$ on raw accuracy by $+1.85\text{ pp}$ ($p = 4.38 \times 10^{-4}$).
  \item Donor rankings also shift with length normalization: on Qwen $F$, cross-installed $F(a_Q)$ attains the highest raw accuracy ($73.67\%$, $+1.25\text{ pp}$ over $F(\mathbf{1})$, $p=0.031$) despite trailing $F(a_F)$ by $>0.050\text{ nats}$ on text NLL; on Llama $J$, swapping $a_S$ into $J$ leaves \texttt{acc\_norm} unchanged ($+0.16\text{ pp}$, $p=0.63$) while reducing raw \texttt{acc} by $-0.76\text{ pp}$ ($[-1.25, -0.33]$, $3$ vs.\ $17$ discordant items, $p = 0.0026$).
\end{itemize}

\begin{table}[!t]
\centering
\caption{Downstream PIQA decision decoupling ($1{,}838$ validation questions) across Qwen3-1.7B precision transfer and Llama-3.2-3B $4$-site context swap. Paired differences report percentage-point shifts, $10{,}000$-draw bootstrap $95\%$ CIs, and exact two-sided McNemar $p$-values.}
\label{tab:piqa_decoupling}
\setlength{\tabcolsep}{5pt}
\renewcommand{\arraystretch}{1.15}
\small
\fittable{%
\begin{tabular}{@{}llrrrr@{}}
\toprule
Family & State / Paired Contrast & Norm Acc (\texttt{acc\_norm}) & Correct & Raw Acc (\texttt{acc}) & Correct \\
\midrule
\multirow{6}{*}{Qwen3-1.7B} & $F(\mathbf{1})$ [$\mathrm{FP}_0$] & $72.31\%$ & $1{,}329$ & $72.42\%$ & $1{,}331$ \\
 & $F(a_F)$ [$\mathrm{FP}_{\mathrm{SELF}}$] & $73.07\%$ & $1{,}343$ & $73.01\%$ & $1{,}342$ \\
 & $F(a_Q)$ [$\mathrm{FP}_{\mathrm{FROM\_Q}}$] & $73.23\%$ & $1{,}346$ & $73.67\%$ & $1{,}354$ \\
 & $Q(\mathbf{1})$ [$Q_0$] & $68.23\%$ & $1{,}254$ & $67.79\%$ & $1{,}246$ \\
 & $Q(a_F)$ [$Q_{\mathrm{FROM\_FP}}$] & $69.26\%$ & $1{,}273$ & $67.63\%$ & $1{,}243$ \\
 & $Q(a_Q)$ [$Q_{\mathrm{SELF}}$] & $70.13\%$ & $1{,}289$ & $69.48\%$ & $1{,}277$ \\
\addlinespace[2pt]
\multirow{3}{*}{\emph{Qwen Contrasts}} & $Q(a_F) - Q(\mathbf{1})$ & \multicolumn{2}{r}{$+1.03\text{ pp}\;[-0.16,\,+2.29]\;(p=0.12)$} & \multicolumn{2}{r}{$-0.16\text{ pp}\;[-1.20,\,+0.87]\;(p=0.84)$} \\
 & $Q(a_Q) - Q(\mathbf{1})$ & \multicolumn{2}{r}{$+1.90\text{ pp}\;[+0.60,\,+3.21]\;(p=0.0055)$} & \multicolumn{2}{r}{$+1.69\text{ pp}\;[+0.44,\,+2.94]\;(p=0.011)$} \\
 & $Q(a_Q) - Q(a_F)$ & \multicolumn{2}{r}{$+0.87\text{ pp}\;[-0.11,\,+1.85]\;(p=0.10)$} & \multicolumn{2}{r}{$+1.85\text{ pp}\;[+0.87,\,+2.88]\;(p=0.00044)$} \\
\midrule
\multirow{2}{*}{\shortstack[l]{Llama-3.2-3B\\($4$-site)}} & $S(a_J) - S(a_S)$ ($74.54\%$ vs.\ $74.59\%$ norm) & \multicolumn{2}{r}{$-0.05\text{ pp}\;[-0.44,\,+0.33]\;(p=1.00)$} & \multicolumn{2}{r}{$+0.38\text{ pp}\;[-0.05,\,+0.82]\;(p=0.14)$} \\
 & $J(a_S) - J(a_J)$ ($75.63\%$ vs.\ $75.46\%$ norm) & \multicolumn{2}{r}{$+0.16\text{ pp}\;[-0.27,\,+0.60]\;(p=0.63)$} & \multicolumn{2}{r}{$-0.76\text{ pp}\;[-1.25,\,-0.33]\;(p=0.0026)$} \\
\bottomrule
\end{tabular}}
\end{table}

\subsection{The 24-Endpoint ARC Readout Audit and Checkpoint Sign Reversal}
\label{sec:act3_readout}

During Qwen3-1.7B calibration ($0 \to 200 \to 600$ updates; constant $\text{lr}=0.01$ for $200$ steps followed by linear decay to $0.001$), the ARC distillation term uses \textsc{Legacy\_Bare}: a plain-text prompt ending in \texttt{Answer:} scored over bare continuations \texttt{"A"}--\texttt{"D"} without a leading space (Appendix~\ref{app:protocol}). Under Qwen's byte-level BPE tokenizer, bare \texttt{"A"} after \texttt{Answer:} receives tiny absolute vocabulary mass on $\mathrm{FP}_0$ ($2.20 \times 10^{-9}$) prior to candidate renormalization.

To test how identical checkpoints behave across tokenization and prompt wrappers, we evaluate all six fixed Qwen checkpoints ($\mathrm{FP}_0, F_{200}, F_{600}\,[=F(a_F)], Q_0, Q_{200}, Q_{600}\,[=Q(a_Q)]$) with \emph{zero new updates} across $569$ held-out ARC validation items under four interfaces ($24$ endpoints total; each compared against $\mathrm{FP}_0$ evaluated under the \emph{same} interface, measuring teacher-distribution KL and top-$1$ agreement with $\mathrm{FP}_0$ rather than ARC gold labels):
\begin{enumerate}[leftmargin=1.6em,itemsep=1pt,topsep=2pt]
  \item \textbf{\textsc{Legacy\_Bare}}: plain prompt ending in \texttt{Answer:}, bare continuations \texttt{"A"}--\texttt{"D"} ($569$ items).
  \item \textbf{\textsc{Space\_Label}}: identical plain prompt, space-prefixed continuations \texttt{" A"}--\texttt{" D"} ($549$ single-token letter items; $20$ numeric \texttt{1}--\texttt{4} items span two tokens).
  \item \textbf{\textsc{Answer\_Text}}: identical plain prompt, multi-token full option text continuations summed in FP32 log-probability.
  \item \textbf{\textsc{Chat\_Label}}: native Qwen3 non-thinking chat template (\texttt{<|im\_start|>user\allowbreak\textbackslash n...\allowbreak<|im\_end|>\allowbreak\textbackslash n<|im\_start|>assistant\allowbreak\textbackslash n<think>\allowbreak\textbackslash n\textbackslash n</think>\allowbreak\textbackslash n\textbackslash n}), scored over bare labels \texttt{"A"}--\texttt{"D"}.
\end{enumerate}

\begin{figure}[!t]
\centering
\includegraphics[width=\linewidth]{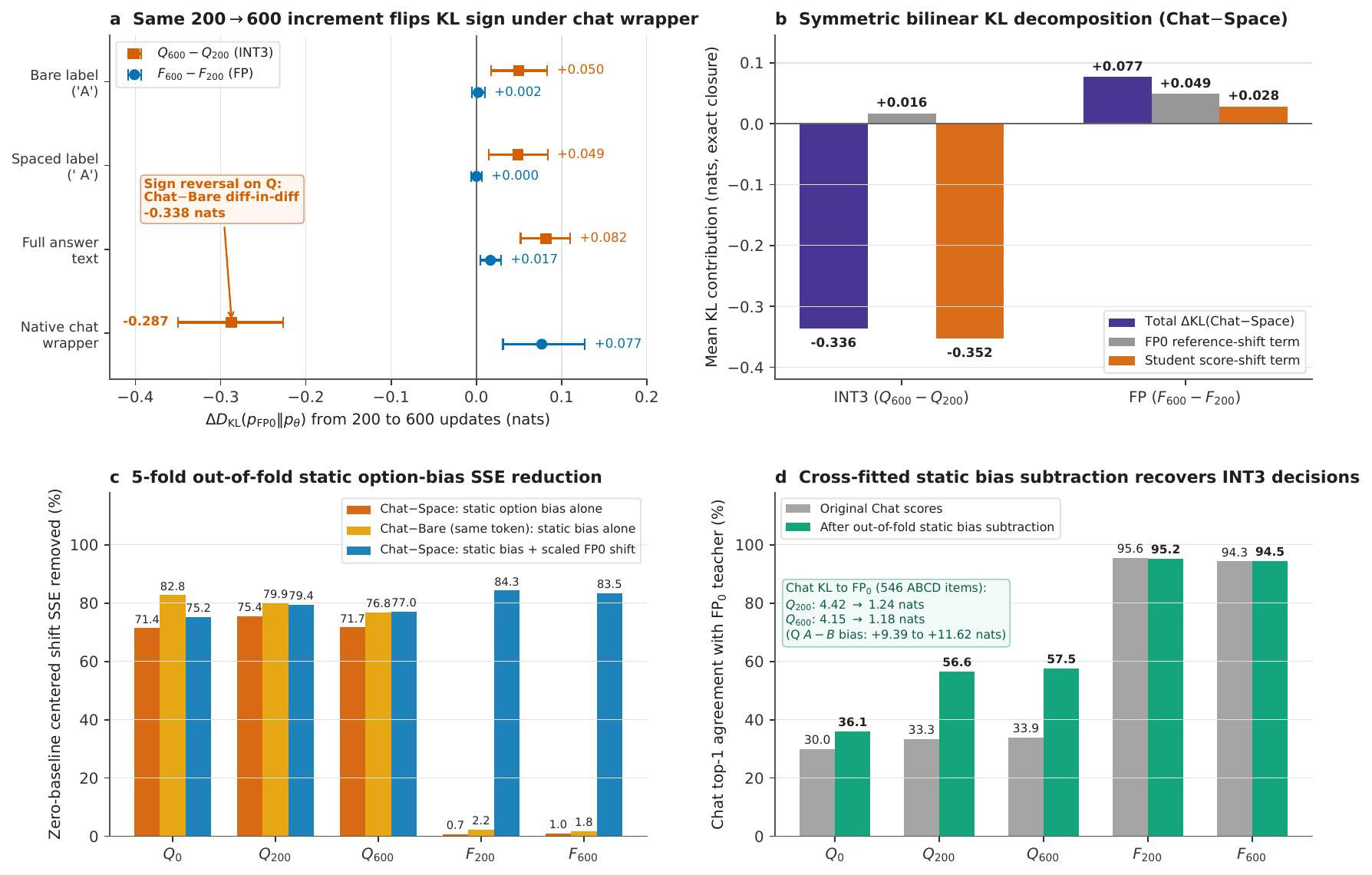}
\caption{\textbf{ARC readout sign reversal, symmetric bilinear KL decomposition, and cross-fitted option-bias subtraction on Qwen3-1.7B.} \textbf{a}, Advancing from $200$ to $600$ updates ($Q_{600} - Q_{200}$) increases INT3 teacher-referenced KL under bare, spaced, and full-text readouts, yet decreases it by $-0.287\text{ nats}$ under native chat wrapping (diff-in-diff $-0.338\text{ nats}$ vs.\ bare). \textbf{b}, Symmetric bilinear KL decomposition~\eqref{eq:bilinear_kl} across all $569$ items ($\text{Chat}-\text{Space}$): student score shifts contribute $-0.352\text{ nats}$ vs.\ $+0.016\text{ nats}$ from $\mathrm{FP}_0$ reference shifts. \textbf{c}, $5$-fold out-of-fold zero-baseline SSE reduction on $546$ \texttt{ABCD} items: static option-label/position bias removes $71.4\%$--$75.4\%$ of INT3 $\text{Chat}-\text{Space}$ shifts and $76.8\%$--$82.8\%$ of same-token $\text{Chat}-\text{Bare}$ shifts vs.\ $<2.3\%$ on FP. \textbf{d}, Post-hoc cross-fitted subtraction of the $4$-parameter static bias lifts INT3 chat $\mathrm{FP}_0$ top-$1$ agreement from $33.3\%$--$33.9\%$ to $56.6\%$--$57.5\%$ and cuts KL by $>2.9\text{ nats}$.}
\label{fig:readout}
\end{figure}

Table~\ref{tab:readout_audit} and Figure~\ref{fig:readout}a establish three results:
\begin{itemize}[leftmargin=1.6em,itemsep=2pt,topsep=2pt]
  \item \textbf{Leading-space tokenization does not alter plain-prompt KL drift.} Switching from \textsc{Legacy\_Bare} to \textsc{Space\_Label} raises $\mathrm{FP}_0$'s single-token candidate mass across the $549$ letter items from $7.76 \times 10^{-11}$ to $\mathbf{0.9997}$, yet the $200 \to 600$ INT3 KL increment is nearly unchanged: $+0.0501\text{ nats}$ ($[+0.0176, +0.0832]$) on \textsc{Legacy\_Bare} versus $+0.0485\text{ nats}$ ($[+0.0148, +0.0836]$) on \textsc{Space\_Label} (diff-in-diff $-0.0015\text{ nats}$ $[-0.0131, +0.0107]$), and $+0.0817\text{ nats}$ on \textsc{Answer\_Text}.
  \item \textbf{Sign reversal under native chat wrapping.} On the exact same $Q_{200}$ and $Q_{600}$ checkpoints, \textsc{Chat\_Label} flips the $200 \to 600$ KL increment to $\mathbf{-0.2875\text{ nats}}$ ($[-0.3497, -0.2267]$; diff-in-diff $-0.3375\text{ nats}$ $[-0.4032, -0.2730]$ vs.\ \textsc{Legacy\_Bare} and $-0.3360\text{ nats}$ vs.\ \textsc{Space\_Label}). Meanwhile, floating-point $F_{600} - F_{200}$ moves in the opposite direction ($+0.0020$ bare and $+0.0001$ spaced vs.\ $+0.0767\text{ nats}$ $[+0.0313, +0.1275]$ in chat).
  \item \textbf{Severe baseline chat distortion and flat teacher agreement.} Despite the $-0.2875\text{ nats}$ late-stage drop, absolute INT3 \textsc{Chat\_Label} divergence from $\mathrm{FP}_0$ remains extreme ($5.6006 \to 4.3364 \to 4.0489\text{ nats}$ KL; top-$1$ agreement with $\mathrm{FP}_0$ stuck at $29.7\% \to 33.7\% \to 34.8\%$, vs.\ $76.6\% \to 77.2\%$ under \textsc{Space\_Label} and $95.8\% \to 94.6\%$ on FP chat; candidate label mass $0.054$--$0.085$ vs.\ $0.961$ on $\mathrm{FP}_0$). Across all four readouts, $Q_{600} - Q_{200}$ top-$1$ teacher-agreement intervals span zero ($-1.58$ to $+1.05\text{ pp}$).
\end{itemize}

\begin{table}[!t]
\centering
\caption{Summary of the $24$-endpoint Qwen3-1.7B ARC readout audit ($569$ validation items; zero new updates). ``Agr.'' denotes top-$1$ choice agreement with the same-interface $\mathrm{FP}_0$ teacher. Complete $24$-endpoint $6$-decimal records appear in Appendix~\ref{app:readout_archival} (Table~\ref{tab:a_readout}).}
\label{tab:readout_audit}
\setlength{\tabcolsep}{4.8pt}
\renewcommand{\arraystretch}{1.15}
\small
\fittable{%
\begin{tabular}{@{}llrrrrrr@{}}
\toprule
Readout Interface & $\mathrm{FP}_0$ Mass$^\dagger$ & $Q_0$ KL & $Q_{200}$ KL & $Q_{600}$ KL & $\Delta\text{KL}_{Q\,(600-200)}$ $[95\%\text{ CI}]$ & $\Delta\text{KL}_{F\,(600-200)}$ & $Q_{600}$ Agr. \\
\midrule
\textsc{Legacy\_Bare} (\texttt{"A"}) & $2.2\times 10^{-9}$ & $1.7375$ & $0.6552$ & $0.7053$ & $+0.0501\;[+0.018,\,+0.083]$ & $+0.0020$ & $75.57\%$ \\
\textsc{Space\_Label} (\texttt{" A"}) & $0.9997$ & $1.2039$ & $0.6539$ & $0.7024$ & $+0.0485\;[+0.015,\,+0.084]$ & $+0.0001$ & $77.15\%$ \\
\textsc{Answer\_Text} (full) & n/a & $1.5198$ & $1.2156$ & $1.2973$ & $+0.0817\;[+0.052,\,+0.110]$ & $+0.0166^*$ & $53.78\%$ \\
\textsc{Chat\_Label} (native) & $0.9608$ & $5.6006$ & $4.3364$ & $4.0489$ & $\mathbf{-0.2875}\;[-0.350,\,-0.227]$ & $+0.0767^*$ & $\mathbf{34.80\%}$ \\
\midrule
\multicolumn{5}{@{}l}{Diff-in-diff ($\textsc{Chat\_Label} - \textsc{Legacy\_Bare}$) on $600\text{u} - 200\text{u}$:} & $-0.3375\;[-0.403,\,-0.273]$ & $+0.0747^*$ & $+2.64\text{ pp}$ \\
\bottomrule
\end{tabular}}
\par\vspace{2pt}\raggedright{\footnotesize $^\dagger$Mean probability mass on single-token candidate labels ($n=569$ for bare/chat; $n=549$ letter items for spaced). $^*95\%$ CI excludes $0$.\par}
\end{table}

\subsection{Bilinear KL Accounting, Option Bias, and Cross-Fitted Subtraction}
\label{sec:act3_mechanism}

Changing from \textsc{Space\_Label} ($s$) to \textsc{Chat\_Label} ($c$) alters both the $\mathrm{FP}_0$ reference distribution $p_m(i)$ and the student's candidate-normalized log-probabilities $\log q_{t,m}(i)$ at step $t \in \{200, 600\}$. Defining the student candidate negative log-probability increment vector $a_m(i) := \log q_{200,m}(i) - \log q_{600,m}(i) \in \mathbb{R}^{|C(x_i)|}$ so that $\Delta\mathrm{KL}_m(i) := D_{\mathrm{KL}}(p_m(i) \| q_{600,m}(i)) - D_{\mathrm{KL}}(p_m(i) \| q_{200,m}(i)) = p_m(i)^\top a_m(i)$, symmetric midpoint accounting with $\bar{p} = \tfrac{1}{2}(p_c + p_s)$ and $\bar{a} = \tfrac{1}{2}(a_c + a_s)$ gives the exact per-item identity
\begin{equation}
  \Delta\mathrm{KL}_c - \Delta\mathrm{KL}_s
  \;=\; \underbrace{(p_c - p_s)^\top \bar{a}}_{\text{Reference-weight shift}}
  \;+\; \underbrace{\bar{p}^\top (a_c - a_s)}_{\text{Student score-shift}}.
  \label{eq:bilinear_kl}
\end{equation}
Note also that because $q_{t,m}(c \mid x_i) = \exp(z_{t,m,c}(i)) / \sum_{j \in C(x_i)} \exp(z_{t,m,j}(i))$, adding any item-level constant to all candidate scores $z_{t,m,\cdot}(i)$ changes full-vocabulary label mass while leaving $q_{t,m}$ and $D_{\mathrm{KL}}(p_m \| q_{t,m})$ algebraically invariant (verified numerically to $1.78 \times 10^{-15}$). Thus conditional KL shifts must arise from relative score differences among candidates rather than total vocabulary label mass. Under the symmetric allocation~\eqref{eq:bilinear_kl} across all $569$ items (closure $\le 8.88 \times 10^{-16}$; Table~\ref{tab:readout_mechanism} Part~A, Figure~\ref{fig:readout}b), INT3's $-0.3360\text{ nats}$ interaction comprises $+0.0164\text{ nats}$ from $(p_c - p_s)^\top \bar{a}$ and $\mathbf{-0.3524\text{ nats}}$ from the student term $\bar{p}^\top (a_c - a_s)$ (versus $+0.0489$ and $+0.0277\text{ nats}$ on FP).

\begin{table}[!t]
\centering
\caption{Readout mechanism and post-hoc bias-subtraction diagnostics on Qwen3-1.7B: $569$-item symmetric bilinear KL accounting~\eqref{eq:bilinear_kl}, $5$-fold cross-fitted zero-baseline SSE reductions~\eqref{eq:sse_red} on $546$ \texttt{ABCD} items (seed $20261017$), and post-hoc out-of-fold static bias subtraction. Full archival values appear in Appendix~\ref{app:readout_archival} (Table~\ref{tab:a_interaction}).}
\label{tab:readout_mechanism}
\setlength{\tabcolsep}{4.5pt}
\renewcommand{\arraystretch}{1.12}
\small
\textit{Part A: Symmetric bilinear KL accounting (569 items; nats)}\par\vspace{3pt}
\fittable{%
\begin{tabular}{@{}lrrrrr@{}}
\toprule
\shortstack[l]{Checkpoint pair\\($600 - 200$)} & $\Delta\mathrm{KL}_{\mathrm{chat}}$ & $\Delta\mathrm{KL}_{\mathrm{space}}$ & \shortstack[r]{Chat$-$Space\\interaction} & \shortstack[r]{$\mathrm{FP}_0$\\reference term} & \shortstack[r]{Student\\shift term} \\
\midrule
INT3 ($Q_{600} - Q_{200}$) & $-0.2875$ & $+0.0485$ & $-0.3360$ & $+0.0164$ & $\mathbf{-0.3524}$ \\
FP ($F_{600} - F_{200}$) & $+0.0767$ & $+0.0001$ & $+0.0766$ & $+0.0489$ & $+0.0277$ \\
\bottomrule
\end{tabular}}
\par\vspace{7pt}
\textit{Part B: Cross-fitted score-shift diagnostics (546 \texttt{ABCD} items)}\par\vspace{3pt}
\fittable{%
\begin{tabular}{@{}lrrrrrr@{}}
\toprule
& \multicolumn{2}{c}{Chat$-$Space SSE red.} & \shortstack[r]{Chat$-$Bare\\SSE red.} & \shortstack[r]{Chat$-$Space\\$A - B$} & \shortstack[r]{Interaction\\SSE red.} & \shortstack[r]{Chat choices\\reproduced} \\
\cmidrule(lr){2-3}
State & Bias & Bias$+\mathrm{FP}_0$ & Bias & (nats) & 2-scale$+$bias & (of 546) \\
\midrule
$Q_0$ & $\mathbf{71.45\%}$ & $75.20\%$ & $\mathbf{82.75\%}$ & $+11.62$ & $89.83\%$ & $507\;(92.9\%)$ \\
$Q_{200}$ & $\mathbf{75.45\%}$ & $79.40\%$ & $\mathbf{79.89\%}$ & $+9.39$ & $93.71\%$ & $504\;(92.3\%)$ \\
$Q_{600}$ & $\mathbf{71.68\%}$ & $77.03\%$ & $\mathbf{76.82\%}$ & $+9.45$ & $93.20\%$ & $494\;(90.5\%)$ \\
$F_{200}$ & $0.67\%$ & $\mathbf{84.31\%}$ & $2.23\%$ & $+0.73$ & $73.35\%$ & $527\;(96.5\%)$ \\
$F_{600}$ & $0.98\%$ & $\mathbf{83.47\%}$ & $1.83\%$ & $+0.10$ & $70.28\%$ & $524\;(96.0\%)$ \\
\addlinespace[1pt]
$\Delta Q_{600-200}$ & $29.42\%$ & $35.35\%$ & --- & $+0.06$ & --- & --- \\
\bottomrule
\end{tabular}}
\par\vspace{7pt}
\textit{Part C: Out-of-fold static bias subtraction (546 \texttt{ABCD} items)}\par\vspace{3pt}
\fittable{%
\begin{tabular}{@{}lrrrrr@{}}
\toprule
& \multicolumn{2}{c}{Chat KL (nats)} & \multicolumn{2}{c}{Teacher agreement (\%)} & \shortstack[r]{Agreement\\change (pp)} \\
\cmidrule(lr){2-3}\cmidrule(lr){4-5}
State & Original & Debiased & Original & Debiased & \\
\midrule
$Q_0$ & $5.5794$ & $2.0599$ & $30.04$ & $36.08$ & $+6.04$ \\
$Q_{200}$ & $4.4187$ & $\mathbf{1.2420}$ & $33.33$ & $\mathbf{56.59}$ & $\mathbf{+23.26}$ \\
$Q_{600}$ & $4.1468$ & $\mathbf{1.1811}$ & $33.88$ & $\mathbf{57.51}$ & $\mathbf{+23.63}$ \\
\addlinespace[2pt]
$F_{200}$ & $0.2361$ & $0.2398$ & $95.60$ & $95.24$ & $-0.37$ \\
$F_{600}$ & $0.3158$ & $0.2960$ & $94.32$ & $94.51$ & $+0.18$ \\
\bottomrule
\end{tabular}}
\par\vspace{3pt}\raggedright{\footnotesize Part A closes to $8.88\times10^{-16}$ per item. Parts B--C use the same fixed five folds; SSE reduction is relative to the zero predictor. These post-hoc CPU diagnostics use archived scores and make zero model calls. Agreement changes are computed before rounding.\par}
\end{table}

To analyze candidate score structure directly, we examine the $546$ standard four-option (\texttt{A,B,C,D}) items using candidate-centered score vectors $\tilde{z}_{M,m}(i) := z_{M,m}(i) - \tfrac{1}{4}\mathbf{1}\mathbf{1}^\top z_{M,m}(i) \in \mathbb{R}^4$. For any target vector $r_i \in \mathbb{R}^4$ and $5$-fold out-of-fold predictor $\hat{r}_i^{(-k(i))}$ (hash seed $20261017$), we measure zero-baseline cross-fitted SSE reduction:
\begin{equation}
  \mathrm{SSE\text{-}Red} := 1 - \frac{\sum_{i=1}^{546} \bigl\|\hat{r}_i^{(-k(i))} - r_i\bigr\|_2^2}{\sum_{i=1}^{546} \|r_i\|_2^2}.
  \label{eq:sse_red}
\end{equation}
Table~\ref{tab:readout_mechanism} (Parts~B--C) and Figure~\ref{fig:readout}c,d establish four mechanistic findings:
\begin{enumerate}[leftmargin=1.6em,itemsep=2pt,topsep=2pt]
  \item \textbf{Dominant static option-label/position bias under quantized chat wrapping.} On each checkpoint's direct wrapper shift $u_M(i) := \tilde{z}_{M,\mathrm{chat}}(i) - \tilde{z}_{M,\mathrm{space}}(i)$, fitting only a constant $4$-vector $\hat{\mu}^{(-k)} \in \mathbb{R}^4$ on four training folds---with zero access to item text---removes $\mathbf{71.45\%}$ ($Q_0$), $\mathbf{75.45\%}$ ($Q_{200}$), and $\mathbf{71.68\%}$ ($Q_{600}$) of zero-baseline SSE on INT3, driven by a $+9.39$ to $+11.62\text{ nats}$ \texttt{A}-over-\texttt{B} log-odds swing (because canonical labels \texttt{A}--\texttt{D} appear in fixed order, this reflects combined label-token and option-position prior bias). On floating point ($F_{200}, F_{600}$), static bias removes only $0.67\%$--$0.98\%$ of SSE, whereas adding a scalar multiple of $\mathrm{FP}_0$'s item shift $\beta u_{\mathrm{FP0}}(i)$ removes $83.47\%$--$84.31\%$ ($\hat{\beta} \approx 1.0$). For the four-term student--teacher interaction $h_M(i) := u_M(i) - u_{\mathrm{FP0}}(i)$, a $5$-parameter two-scale-plus-bias model removes $89.83\%$--$93.71\%$ of SSE on INT3 and reproduces $90.5\%$--$92.9\%$ of quantized chat argmax choices.
  \item \textbf{Same-output-token control ($\text{Chat}-\text{Bare}$).} Because $\text{Chat}-\text{Space}$ changes both the prompt wrapper and the leading space on candidate tokens, we run a post-hoc CPU control on $\tilde{z}_{M,\mathrm{chat}}(i) - \tilde{z}_{M,\mathrm{bare}}(i)$, where both interfaces score the \emph{exact same} bare tokens \texttt{"A"}--\texttt{"D"}. Out-of-fold static option bias still removes $\mathbf{82.75\%}$ ($Q_0$, $\texttt{A}-\texttt{B} = +15.32\text{ nats}$), $\mathbf{79.89\%}$ ($Q_{200}$, $+10.18\text{ nats}$), and $\mathbf{76.82\%}$ ($Q_{600}$, $+10.35\text{ nats}$) of zero-baseline SSE versus $2.23\%$ ($F_{200}$) and $1.83\%$ ($F_{600}$), confirming that the static bias arises from the chat wrapper under INT3 quantization rather than candidate-token spacing.
  \item \textbf{Direct decision recovery via cross-fitted static bias subtraction.} To test whether this static bias directly causes INT3's chat decision collapse, a post-hoc CPU intervention subtracts each fold's training-estimated $4$-vector mean of $u_M$ from held-out \textsc{Chat\_Label} scores $z_{M,\mathrm{chat}}(i)$ ($546$ \texttt{ABCD} items, zero gold labels, zero hyperparameter tuning; Table~\ref{tab:readout_mechanism} Part~C, Figure~\ref{fig:readout}d). Subtracting those four numbers slashes held-out INT3 chat KL to $\mathrm{FP}_0$ from $5.5794 \to 2.0599$ ($Q_0$), $\mathbf{4.4187 \to 1.2420}$ ($Q_{200}$), and $\mathbf{4.1468 \to 1.1811}$ ($Q_{600}$), while raising held-out $\mathrm{FP}_0$ teacher top-$1$ agreement from $30.04\% \to 36.08\%$ ($Q_0$), $\mathbf{33.33\% \to 56.59\%}$ ($Q_{200}$), and $\mathbf{33.88\% \to 57.51\%}$ ($Q_{600}$); floating-point agreement is virtually unchanged ($95.60\% \to 95.24\%$ on $F_{200}$; $94.32\% \to 94.51\%$ on $F_{600}$).
  \item \textbf{State-level static bias vs.\ unexplained late-stage item variation.} Whereas static option bias dominates each individual INT3 checkpoint's wrapper distortion ($>71\%$), applying the same $5$-fold fit to the $200 \to 600$ \emph{increment} $u_{Q_{600}}(i) - u_{Q_{200}}(i)$ removes $29.42\%$ of zero-baseline SSE with static bias ($35.35\%$ with linear $\mathrm{FP}_0$ shift), leaving $64.65\%$ as item-dependent score-shift variation not captured by those two low-dimensional predictors.
\end{enumerate}

\section{Related Work}
\label{sec:related}

\paragraph{Post-training quantization, scale optimization, and cross-configuration reuse.}
Sub-$4$-bit LLM PTQ combines second-order weight rounding \citep{nagel2020adaround,li2021brecq,frantar2023gptq,kim2025guidedquant} with multiplicative rescaling \citep{xiao2023smoothquant,lin2024awq,shao2024omniquant}. Post-discretization scale optimization is well established: PEQA \citep{kim2023peqa} fine-tuned quantization scales alone for downstream adaptation with frozen integer codes, and EfficientQAT \citep{chen2025efficientqat} introduced end-to-end step-size training (\texttt{E2E-QP}) after blockwise reconstruction. For cross-configuration reuse, CoA-LoRA \citep{ye2026coalora} trains a hypernetwork to generate quantization-conditioned LoRA adapters, whereas Task Arithmetic \citep{ilharco2023editing} and $\text{(IA)}^3$ \citep{liu2022fewshot} manipulate additive weight deltas or activation multipliers in floating point. Rather than proposing scale training as a new paradigm or learning an adapter generator, our contribution is to test direct native installation ($s_g \leftarrow a_g s_{g,0}$) of a single relative gain vector across FP and INT3 representations and to isolate recipient-source preference.

\paragraph{Cross-layer PTQ interactions and small-calibration-set overfitting.}
Prior work has shown that blockwise PTQ errors interact non-additively across layers \citep{shabanovi2024interactions,deng2025clado,noda2026crosslayer}, and \citet{hubara2021accurate} analyzed how optimizing quantized parameters on small calibration sets induces overfitting that motivates blockwise objectives. Our Llama-3.2-3B $S \leftrightarrow J$ control adds a sharp causal identification of non-local context sensitivity by holding all calibrated initial tensors bit-identical ($\delta_g^{(S,J)} \equiv 0$). Complementing EfficientQAT \citep{chen2025efficientqat}---which optimizes direct step sizes rather than log-scales, so raw learning-rate magnitudes are not directly interchangeable across parameterizations---our $4$-site versus $196$-site ($22.02\text{M}$-scale) experiments quantify consumed-loss pilot failure, base-dependent PIQA degradation, and asynchronous multi-loss trajectories on small PTQ pools.

\paragraph{Quantization confidence distortion and prompt/label prior biases.}
Prompt sensitivity and static option/label prior biases in language-model multiple-choice evaluation were established by \citet{zhao2021calibrate}, who corrected output probabilities by subtracting estimated prior biases, while \citet{proskurina2024quantization} showed that low-bit quantization distorts model confidence nonuniformly across inputs. Our $24$-endpoint Qwen3-1.7B readout audit links these two lines directly to scale calibration: identical $200 \to 600$ checkpoint updates invert teacher-KL rankings between plain and chat prompts, and out-of-fold subtraction of a $4$-parameter quantized chat option bias recovers $>23\text{ pp}$ of lost teacher agreement without touching floating-point checkpoints.

\section{Discussion, Scope, and Practical Guidelines}
\label{sec:discussion}

Across Qwen3-1.7B and Llama-3.2-3B, three empirical principles emerge for post-training scale calibration under composite objectives:
\begin{enumerate}[leftmargin=1.6em,itemsep=2pt,topsep=2pt]
  \item \textbf{Export calibration updates as relative groupwise gains.} Storing $a_g = s_{g,\star}/s_{g,0}$ in canonical input-column group coordinates allows immediate zero-training installation across floating-point and INT3 checkpoints sharing the same group layout, retaining $77\%$--$91\%$ of self-calibration text-NLL reductions across precisions while leaving resolved own-source and non-local context preferences when recipient-specific refinement is affordable.
  \item \textbf{Guard against small-pool memorization when unlocking network-wide scales.} On small PTQ calibration sets ($\le 256$ sequences), unlocking all linear projections ($22\text{M}$ log-scales on $3\text{B}$ models) makes consumed calibration loss an unreliable learning-rate selector: aggressive step sizes ($\text{lr}=10^{-2}$ in our log-scale Adam setup) memorize repeated calibration blocks while inflating unseen perplexity $>2\times$ over $4$-site calibration. Conservative step sizes ($\text{lr}=10^{-3}$ here) and held-out per-domain validation checks are required.
  \item \textbf{Audit low-bit checkpoints under deployment prompt wrappers and check static option bias.} Plain-prompt and chat-wrapped teacher KL can move in opposite directions across identical checkpoints. When evaluating quantized models on multiple-choice prompts under chat templates, check candidate-centered option priors (and out-of-fold prior subtraction \citep{zhao2021calibrate}) alongside discrete decision agreement rather than relying on a single uncalibrated KL readout.
\end{enumerate}

\paragraph{Scope boundaries.}
Our results cover Qwen3-1.7B and Llama-3.2-3B under asymmetric INT3 group-$128$ quantization, fixed composite calibration objectives, and single-seed deterministic trajectories (so bootstrap intervals quantify evaluation-sample uncertainty for fixed trained artifacts rather than across-seed training variance). Fresh Qwen confirmation cohorts exclude all project-consumed texts, though public corpora may still overlap foundation-model pretraining data. In §\ref{sec:act2}, comparing $196$-site $\text{lr}=10^{-3}$ against $4$-site $\text{lr}=10^{-2}$ reflects practical operating-point replacement, whereas single-factor capacity and step-size contrasts are isolated separately in Table~\ref{tab:llama_capacity}. In §\ref{sec:act3}, the ARC readout audit evaluates alignment with the matched-interface $\mathrm{FP}_0$ teacher; ground-truth task accuracy is evaluated separately on PIQA. Fixed-order \texttt{ABCD} bias combines token identity with option position.

\FloatBarrier

\clearpage
\appendix
\makeatletter
\setlength{\@fptop}{0pt}
\setlength{\@fpsep}{12pt}
\makeatother
\setcounter{figure}{0}
\setcounter{table}{0}
\renewcommand{\thefigure}{A\arabic{figure}}
\renewcommand{\thetable}{A\arabic{table}}

\section{Exact Numerical Protocols, Data Isolation, and Execution Recipe}
\label{app:protocol}

\subsection{Native Scale Absorption and CPU Payload Timing}
The gain installer operates on saved FP32 floating-point multipliers and native quantized scale tensors. Quantized donor terminal and initial scales $(s_{g,\star}, s_{g,0})$ are promoted to IEEE-754 FP64 for ratio extraction $a_g = s_{g,\star}/s_{g,0}$; floating-point donor multipliers are promoted to FP64 directly. Recipient installation multiplies in FP64 and casts once to FP32 upon writeback, verifying positivity, finiteness, unchanged integer codes and zero points, and canonical input-column group alignment. Table~\ref{tab:a_cpu} reports single-invocation CPU update-payload construction timings on an AMD Ryzen~5 5600GT ($6$ PyTorch threads, Windows~11, PyTorch 2.13.0+cpu, NumPy 2.5.2) with warm OS cache (shared source load $0.347\text{ s}$, FP64 extraction $0.011\text{ s}$).

\begin{table}[!htbp]
\centering
\caption{Single-invocation CPU costs for constructing Qwen3-1.7B four-projection cross-transfer update payloads ($393{,}216$ groups; $1{,}572{,}864\text{ bytes}$ in FP32). Identity checking on $Q$ verifies unchanged surrounding native quantized projections.}
\label{tab:a_cpu}
\setlength{\tabcolsep}{7pt}
\renewcommand{\arraystretch}{1.14}
\small
\fittable{%
\begin{tabular}{@{}lrrr@{}}
\toprule
Installed Recipient Payload & Arithmetic (ms) & Identity Checks (ms) & Write \& Flush (ms) \\
\midrule
$F$ with $Q$-gain ($4$ dense FP32 matrices) & $137.45$ & $226.92$ & $1{,}058.01$ \\
$Q$ with $F$-gain ($4$ native scale tensors) & $\mathbf{3.27}$ & $760.38$ & $112.77$ \\
\bottomrule
\end{tabular}}
\end{table}

\subsection{Calibration Schedules, Data Isolation, and Statistical Estimators}
\paragraph{Qwen3-1.7B ($200 + 400$ two-phase schedule).} Weights and tokenizer come from \nolinkurl{Qwen/Qwen3-1.7B} (commit \texttt{70d244cc\allowbreak 86ccca08\allowbreak cf5af4e1\allowbreak e306ecf9\allowbreak 08b1ad5e}). Native $Q_0$ uses asymmetric INT3, group size $128$, GPTQ block size $128$, Hessian damping $0.01$, \texttt{act\_order=True} remapped to canonical input-column groups, and no MSE clipping. Both $F$ and $Q$ donors optimize $393{,}216$ log-scales across $0$-based layers $6, 13, 20, 27$ \nolinkurl{mlp.down_proj} for $600$ Adam steps ($\beta=(0.9,0.999)$, $\epsilon=10^{-8}$, zero weight decay, deterministic cyclic batch cursors seeded with $20261012$), consuming $8$ ARC items, $1$ WikiText block ($2{,}048$ tokens; $2{,}047$ scored next-token positions), and $1$ C4 block ($2{,}048$ tokens) per step with fixed objective weights $(0.0806241550, 0.4067881928, 0.5125876522)$ computed once on $Q_0$ as inverse initial \textsc{Fit} gradient $\ell_2$ norms and reused identically on $F$. Both $F$ and $Q$ execute the exact same two-phase learning-rate schedule:
\begin{itemize}[leftmargin=1.6em,itemsep=1pt,topsep=2pt]
  \item \textbf{Phase 1 (steps $t = 1,\dots,200$, \texttt{COMMON\_200})}: constant $\text{lr}(t) = 0.01$, producing $F_{200}$ and $Q_{200}$.
  \item \textbf{Phase 2 (steps $t = 201,\dots,600$, \texttt{LINEAR\_DECAY})}: continuing from the exact $t=200$ parameters, Adam moments, and data cursors with pre-step learning rate $\text{lr}(t) = 0.01 - 0.009(t - 200)/400$ ($\text{lr}(201)=0.0099775$, $\text{lr}(400)=0.0055$, $\text{lr}(600)=0.001$; terminal cursors at step $600$: ARC epoch $8$ index $64$, WikiText epoch $2$ index $72$, C4 epoch $2$ index $80$), producing $F_{600}$ ($\mathrm{FP}_{\mathrm{SELF}}$) and $Q_{600}$ ($Q_{\mathrm{SELF}}$).
\end{itemize}
As summarized in Table~\ref{tab:a_data_roles}, Qwen donor training consumed $264$ WikiText blocks ($184$ articles), while common \textsc{Fit} diagnostics used a $29$-block WikiText subset alongside the $260$ C4 blocks and $592$ ARC items. Fresh confirmation excludes all $932$ project-consumed WikiText article IDs plus official validation/test splits ($1{,}054$ normalized titles/hashes; salt \nolinkurl{calibration-transfer-confirmation-20261014} selects $139$ unseen articles / $256$ blocks from raw WikiText-103 \texttt{b08601e0}) and all $319{,}032$ C4 documents in shards $0$--$6$ ($1{,}103$ unseen documents / $256$ blocks from C4 validation shard $7$, revision \texttt{1588ec45}; scored in FP32/SDPA with TF32 disabled on one RTX~5090 in $704.872\text{ s}$).

\begin{table}[!htbp]
\centering
\caption{Data-role isolation ledger across Qwen3-1.7B and Llama-3.2-3B experiments. Every text block contains $2{,}048$ contiguous tokens; exact token-block and article/document intersections between calibration training and held-out evaluation splits are empty.}
\label{tab:a_data_roles}
\setlength{\tabcolsep}{5.5pt}
\renewcommand{\arraystretch}{1.14}
\small
\fittable{%
\begin{tabular}{@{}llrrrr@{}}
\toprule
Model Family & Split / Experimental Role & WikiText Blocks & C4 Blocks & ARC Items & PIQA Items \\
\midrule
\multirow{5}{*}{Qwen3-1.7B} & Donor calibration training ($600\text{u}$) & $264$ ($184$ art.) & $260$ ($1{,}244$ docs) & $592$ & $0$ \\
 & Common \textsc{Fit} diagnostic subset & $29$ & $260$ & $592$ & $0$ \\
 & Development: \textsc{Validation} ($113$--$114$ art.) & $255$ & $260$ ($1{,}128$ docs) & $569$ & $1{,}838$ \\
 & Development: \textsc{Replay} ($121$ art.) & $255$ & $261$ ($1{,}145$ docs) & n/a & n/a \\
 & \textbf{Fresh unseen confirmation} & $\mathbf{256}$ ($139$ art.) & $\mathbf{256}$ ($1{,}103$ docs) & n/a & n/a \\
\midrule
\multirow{2}{*}{Llama-3.2-3B} & Calibration training \& \textsc{Fit} ($50\text{u}/200\text{u}$) & $28$ & $256$ & $592$ & $0$ \\
 & Held-out \textsc{Validation} ($114$ art.) & $256$ & $256$ ($1{,}128$ docs) & $569$ & $1{,}838$ \\
\bottomrule
\end{tabular}}
\end{table}

\paragraph{Llama-3.2-3B and Resampling Estimators.} Llama-3.2-3B runs ($4$-site $786{,}432$ scales and $196$-site $22{,}020{,}096$ scales; base weight and tokenizer file SHA-256 digests recorded in \texttt{revision\_evidence.json}) use sampler seed $20261002$, Adam $\beta=(0.9,0.999)$, $\epsilon=10^{-8}$, zero weight decay, constant learning rates $\text{lr} \in \{0.01, 0.001\}$, and fixed objective weights $(0.0265874055, 0.3424030653, 0.6310095291)$ on the $592$-ARC / $28$-WikiText / $256$-C4 \textsc{Fit} pool. All four $4$-site crossed states ($S_{\mathrm{SELF}}, S_{\mathrm{FROM\_J}}, J_{\mathrm{SELF}}, J_{\mathrm{FROM\_S}}$) were re-evaluated on one RTX~5090 ($995.054\text{ s}$), matching archived loss vectors within $\le 9.22 \times 10^{-7}$ mean WikiText NLL, $5.89 \times 10^{-8}$ C4 NLL, and $2.30 \times 10^{-7}$ ARC KL. Paired bootstrap intervals use $10{,}000$ draws (seeds $20261014$ for Qwen/PIQA/capacity/C4-block sensitivity, $20261004$ for Llama $4$-site, $20261019$ for placement, $20261016$ for ARC readout), resampling WikiText by article cluster weighted by retained block count, C4 by paired block, and ARC/PIQA by paired item. Plain-text ARC prompts use \texttt{add\_special\_tokens=True} (prepending BOS \texttt{128000} on Llama; no BOS on Qwen) and continuations use \texttt{add\_special\_tokens=False}, computing $p_\theta(c \mid x) = \exp(z_c) / \sum_{j \in C(x)} \exp(z_j)$ from unnormalized FP32 sequence log-probabilities $z_c$.

\subsection{Reproducibility Scope and From-Scratch Execution Recipe}
\label{app:recipe}
Two distinct reproducibility tiers are provided in the accompanying supplement (\nolinkurl{calibration-transfer-artifact.zip}):
\begin{enumerate}[leftmargin=1.6em,itemsep=1pt,topsep=2pt]
  \item \textbf{Weight-free CPU statistical and score-level reproduction.} Running \nolinkurl{reproduce.py}, \texttt{install\_gain.py -{}-self-check}, and \nolinkurl{plot_figures.py} in isolated Python (\texttt{python -I}) with zero model weights or GPU calls recomputes all $14$ unit-level cohorts, the $24$-endpoint ARC readout audit, the $5$-fold cross-fitted regressions and post-hoc bias-subtraction / same-token controls, C4 span-audit sensitivities (\nolinkurl{audit_c4.py}), and all four vector figures to within $2 \times 10^{-11}$ of archived values.
  \item \textbf{From-scratch checkpoint optimization and forward scoring.} Regenerating donor checkpoints and raw unit journals from upstream \texttt{Qwen/\allowbreak Qwen3-1.7B} (\texttt{70d244cc}) and \texttt{meta-llama/\allowbreak Llama-3.2-3B} weights follows the step-by-step recipe in \texttt{RECIPE.md} (bundled alongside the \texttt{manifests/} files \texttt{qwen\_\allowbreak training.json}, \texttt{llama\_\allowbreak configurations.csv}, and \texttt{qwen\_\allowbreak fresh\_\allowbreak wiki.json}, \texttt{qwen\_\allowbreak fresh\_\allowbreak c4.json}, and \texttt{qwen\_\allowbreak fresh\_\allowbreak c4\_\allowbreak spans.json}): (i) quantize projections via GPTQ or RTN per \texttt{llama\_\allowbreak configurations.csv} and remap \texttt{act\_\allowbreak order} columns to canonical group indices; (ii) run exact FP32/SDPA log-scale Adam optimization under the $200+400$ Qwen schedule or constant-LR Llama grid; (iii) extract and cross-install FP64-promoted relative gains via \texttt{install\_\allowbreak gain.py}; and (iv) score held-out $2{,}048$-token text blocks, PIQA completions, and the four ARC prompt interfaces.
\end{enumerate}

\section{Archival 6-Decimal Ledgers for Qwen3-1.7B}
\label{app:qwen_archival}

Tables~\ref{tab:a_qwen_means}--\ref{tab:a_placement} record complete $6$-decimal archival means, representation gaps, development-split paired contrasts (\textsc{Replay} and \textsc{Validation}), within-projection permutation means, and all $16$ norm-matched layer-withdrawal contrasts reproduced by \texttt{reproduce.py}.

\begin{table}[!htbp]
\centering
\caption{Complete $6$-decimal Qwen3-1.7B state means, perplexities, and precision gaps $G(a)$ across Fresh Confirmation and \textsc{Replay} development cohorts (PIQA correct counts out of $1{,}838$ appear in Table~\ref{tab:piqa_decoupling}).}
\label{tab:a_qwen_means}
\setlength{\tabcolsep}{4.8pt}
\renewcommand{\arraystretch}{1.14}
\footnotesize
\fittable{%
\begin{tabular}{@{}llrrrrrr@{}}
\toprule
State / Gap & Gain & Fresh Wiki NLL & Fresh Wiki PPL & Fresh C4 NLL & Fresh C4 PPL & Replay Wiki NLL & Replay C4 NLL \\
\midrule
$\mathrm{FP}_0$ [$F(\mathbf{1})$] & None & $2.815478$ & $16.701156$ & $3.128175$ & $22.832262$ & $2.868635$ & $3.031399$ \\
$\mathrm{FP}_{\mathrm{SELF}}$ [$F(a_F)$] & $a_F$ & $2.308516$ & $10.059482$ & $2.860485$ & $17.469995$ & $2.350632$ & $2.765396$ \\
$\mathrm{FP}_{\mathrm{FROM\_Q}}$ [$F(a_Q)$] & $a_Q$ & $2.358858$ & $10.578860$ & $2.916725$ & $18.480667$ & $2.401364$ & $2.818406$ \\
$Q_0$ [$Q(\mathbf{1})$] & None & $3.099833$ & $22.194237$ & $3.405603$ & $30.132461$ & $3.163441$ & $3.286023$ \\
$Q_{\mathrm{FROM\_FP}}$ [$Q(a_F)$] & $a_F$ & $2.548046$ & $12.782105$ & $3.162151$ & $23.621359$ & $2.601065$ & $3.040561$ \\
$Q_{\mathrm{SELF}}$ [$Q(a_Q)$] & $a_Q$ & $2.490195$ & $12.063631$ & $3.088998$ & $21.955076$ & $2.539240$ & $2.971454$ \\
\midrule
Gap $G(\mathbf{1})$ & None & $0.284355$ & --- & $0.277428$ & --- & $0.294806$ & $0.254624$ \\
Gap $G(a_F)$ & $a_F$ & $0.239530$ & --- & $0.301667$ & --- & $0.250434$ & $0.275165$ \\
Gap $G(a_Q)$ & $a_Q$ & $0.131338$ & --- & $0.172273$ & --- & $0.137876$ & $0.153047$ \\
\bottomrule
\end{tabular}}
\end{table}

\begin{table}[!htbp]
\centering
\caption{Archival $6$-decimal paired NLL contrasts and $95\%$ bootstrap CIs across Fresh Confirmation, \textsc{Replay} development, \textsc{Validation} development, and \textsc{Replay} group-permutation cohorts on Qwen3-1.7B.}
\label{tab:a_qwen_dev}
\setlength{\tabcolsep}{5pt}
\renewcommand{\arraystretch}{1.14}
\footnotesize
\fittable{%
\begin{tabular}{@{}llrr@{}}
\toprule
Cohort & Paired Contrast & WikiText $\Delta$NLL $[95\%\text{ CI}]$ & C4 $\Delta$NLL $[95\%\text{ CI}]$ \\
\midrule
\multirow{7}{*}{Fresh ($256$/$256$)} & $Q(a_F) - Q(\mathbf{1})$ & $-0.551787\;[-0.564876,\,-0.538694]$ & $-0.243452\;[-0.250417,\,-0.236242]$ \\
 & $F(a_Q) - F(\mathbf{1})$ & $-0.456620\;[-0.465492,\,-0.447597]$ & $-0.211449\;[-0.217559,\,-0.205294]$ \\
 & $Q(a_Q) - Q(\mathbf{1})$ & $-0.609637\;[-0.623438,\,-0.595710]$ & $-0.316605\;[-0.324400,\,-0.308563]$ \\
 & $F(a_F) - F(\mathbf{1})$ & $-0.506962\;[-0.516455,\,-0.497389]$ & $-0.267690\;[-0.274584,\,-0.260718]$ \\
 & $d_Q = Q(a_Q) - Q(a_F)$ & $-0.057851\;[-0.059645,\,-0.055978]$ & $-0.073153\;[-0.075296,\,-0.070911]$ \\
 & $d_F = F(a_Q) - F(a_F)$ & $+0.050342\;[+0.048949,\,+0.051781]$ & $+0.056240\;[+0.054000,\,+0.058631]$ \\
 & Interaction $\mathcal{I} = d_Q - d_F$ & $-0.108193\;[-0.110546,\,-0.105810]$ & $-0.129393\;[-0.133285,\,-0.125615]$ \\
\midrule
\multirow{5}{*}{\textsc{Replay} ($255$/$261$)} & $Q(a_F) - Q(\mathbf{1})$ & $-0.562376\;[-0.579042,\,-0.545152]$ & $-0.245462\;[-0.253218,\,-0.237597]$ \\
 & $F(a_Q) - F(\mathbf{1})$ & $-0.467271\;[-0.481330,\,-0.452876]$ & $-0.212993\;[-0.220396,\,-0.205271]$ \\
 & $d_Q = Q(a_Q) - Q(a_F)$ & $-0.061825\;[-0.063898,\,-0.059788]$ & $-0.069108\;[-0.071549,\,-0.066570]$ \\
 & $d_F = F(a_Q) - F(a_F)$ & $+0.050733\;[+0.049149,\,+0.052427]$ & $+0.053011\;[+0.051368,\,+0.054660]$ \\
 & Interaction $\mathcal{I} = d_Q - d_F$ & $-0.112558\;[-0.114904,\,-0.110072]$ & $-0.122118\;[-0.125586,\,-0.118473]$ \\
\midrule
\multirow{3}{*}{\textsc{Validation} ($255$/$260$)} & $Q(a_F) - Q(\mathbf{1})$ & $-0.563930\;[-0.583677,\,-0.540339]$ & $-0.247157\;[-0.254107,\,-0.240066]$ \\
 & $F(a_Q) - F(\mathbf{1})$ & $-0.460428\;[-0.476524,\,-0.441403]$ & $-0.219395\;[-0.225537,\,-0.213125]$ \\
 & Interaction $\mathcal{I} = d_Q - d_F$ & $-0.112485\;[-0.115790,\,-0.109242]$ & $-0.126137\;[-0.129612,\,-0.122645]$ \\
\midrule
\multirow{2}{*}{Shuffle (\textsc{Replay})} & $Q(a_F)$: $\text{Shuffled} - \text{Aligned}$ & $+0.595840\;[+0.576709,\,+0.614823]$ & $+0.282141\;[+0.273744,\,+0.290410]$ \\
 & $F(a_Q)$: $\text{Shuffled} - \text{Aligned}$ & $+0.485449\;[+0.470579,\,+0.500156]$ & $+0.223070\;[+0.215348,\,+0.230536]$ \\
\bottomrule
\end{tabular}}
\end{table}

\begin{table}[!htbp]
\centering
\caption{Complete $16$-cell archival $6$-decimal ledger for norm-matched layer withdrawal ($\text{DROP} - \text{UNIFORM}$ at $200$ updates on Qwen3-1.7B \textsc{Validation}).}
\label{tab:a_placement}
\setlength{\tabcolsep}{5.5pt}
\renewcommand{\arraystretch}{1.22}
\footnotesize
\fittable{%
\begin{tabular}{@{}lrrlrr@{}}
\toprule
$Q_{200}$ Layer & WikiText $\Delta$NLL $[95\%\text{ CI}]$ & C4 $\Delta$NLL $[95\%\text{ CI}]$ & $F_{200}$ Layer & WikiText $\Delta$NLL $[95\%\text{ CI}]$ & C4 $\Delta$NLL $[95\%\text{ CI}]$ \\
\midrule
Layer $6$ & \ci{$+0.056555$}{$[0.054327,\,0.058634]$} & \ci{$+0.031709$}{$[0.030314,\,0.033134]$} & Layer $6$ & \ci{$+0.051238$}{$[0.049373,\,0.052974]$} & \ci{$+0.032266$}{$[0.030678,\,0.033929]$} \\
\addlinespace[2pt]
Layer $13$ & \ci{$+0.016099$}{$[0.015150,\,0.017093]$} & \ci{$+0.012712$}{$[0.012054,\,0.013409]$} & Layer $13$ & \ci{$+0.011932$}{$[0.010948,\,0.012830]$} & \ci{$+0.007916$}{$[0.007475,\,0.008366]$} \\
\addlinespace[2pt]
Layer $20$ & \ci{$+0.037774$}{$[0.036441,\,0.039228]$} & \ci{$+0.013725$}{$[0.013116,\,0.014313]$} & Layer $20$ & \ci{$+0.025438$}{$[0.024403,\,0.026428]$} & \ci{$+0.006327$}{$[0.005907,\,0.006733]$} \\
\addlinespace[2pt]
Layer $27$ & \ci{$+0.080062$}{$[0.077308,\,0.082660]$} & \ci{$+0.073418$}{$[0.070684,\,0.076128]$} & Layer $27$ & \ci{$+0.072699$}{$[0.070484,\,0.074822]$} & \ci{$+0.066214$}{$[0.063893,\,0.068567]$} \\
\bottomrule
\end{tabular}}
\end{table}

\clearpage
\section{Llama-3.2-3B Non-Local Context Manifest and Block-Span Sensitivity}
\label{app:llama_context}

Tables~\ref{tab:a_llama_sites} and~\ref{tab:a_c4_block} document the exact $9$ uncalibrated projection differences between $S$ (\textsc{Static31}) and $J$ (\textsc{Joint39}) alongside $6$-decimal Llama $4$-site means/contrasts and circular moving-block C4 bootstrap intervals across segment lengths $L \in \{1,2,4,8,16\}$.

\begin{table}[!htbp]
\centering
\caption{Part A: All nine encoding differences between Llama-3.2-3B INT3 configurations $S$ and $J$ across the $196$ linear projections (flat projection indices $28, 50, 56, 57, 89, 90, 103, 110, 121$; all four calibrated sites at layers $6, 13, 20, 27$ \texttt{mlp.down\_proj} are bit-identical). Part B: $6$-decimal archival means and paired contrasts (seed $20261004$).}
\label{tab:a_llama_sites}
\setlength{\tabcolsep}{4.8pt}
\renewcommand{\arraystretch}{1.16}
\footnotesize
\fittable{%
\begin{tabular}{@{}rllllr@{}}
\toprule
\multicolumn{6}{@{}l}{\emph{Part A: Complete $9$-Projection External Encoding Differences ($0$-based layer indices; INT3 group size $128$)}} \\
\midrule
Flat Index & Layer & Projection & $S$ (\textsc{Static31}) & $J$ (\textsc{Joint39}) & Calibrated? \\
\midrule
$28$ / $56$ & $4$ / $8$ & \texttt{self\_attn.q\_proj} & GPTQ ($1\%$ damp.) & GPTQ ($0.1\%$ damp.) & No \\
$50$ & $7$ & \texttt{self\_attn.k\_proj} & GPTQ ($0.1\%$ damp.) & Round-to-nearest (RTN) & No \\
$57$ & $8$ & \texttt{self\_attn.k\_proj} & GPTQ ($1\%$ damp.) & GPTQ ($0.1\%$ damp.) & No \\
$89$ / $103$ & $12$ / $14$ & \texttt{mlp.up\_proj} & GPTQ ($1\%$ damp.) & GPTQ ($0.1\%$ damp.) & No \\
$90$ & $12$ & \texttt{mlp.down\_proj} & GPTQ ($1\%$ damp.) & GPTQ ($0.1\%$ damp.) & No \\
$110$ / $121$ & $15$ / $17$ & \texttt{mlp.up\_proj} / \texttt{v\_proj} & GPTQ ($1\%$ damp.) & Round-to-nearest (RTN) & No \\
\midrule
\multicolumn{6}{@{}l}{\emph{Part B: $6$-Decimal $4$-Site Llama-3.2-3B State Means and Paired Cross-Minus-Own Contrasts}} \\
\midrule
\multicolumn{2}{@{}l}{State / Contrast} & ARC KL & WikiText NLL & C4 NLL & PIQA Norm / Raw \\
\midrule
\multicolumn{2}{@{}l}{$Q_0$ own gain ($Q_{\mathrm{FOUR}}$, ref.)} & $0.102966$ & $2.248734$ & $2.706076$ & $1{,}363$ / $1{,}352$ \\
\multicolumn{2}{@{}l}{$S_0$ / $S_{\mathrm{SELF}}$ / $S_{\mathrm{FROM\_J}}$} & \ci{$0.177394$ / $0.100022$}{$0.106661$ (cross)} & \ci{$2.369735$ / $2.263367$}{$2.264672$ (cross)} & \ci{$2.786582$ / $2.705577$}{$2.706699$ (cross)} & \ci{$1{,}371$/$1{,}369$}{$1{,}370$/$1{,}376$} \\
\addlinespace[2pt]
\multicolumn{2}{@{}l}{$J_0$ / $J_{\mathrm{SELF}}$ / $J_{\mathrm{FROM\_S}}$} & \ci{$0.173054$ / $0.105952$}{$0.109944$ (cross)} & \ci{$2.369435$ / $2.264572$}{$2.265354$ (cross)} & \ci{$2.785321$ / $2.704759$}{$2.705872$ (cross)} & \ci{$1{,}387$/$1{,}371$}{$1{,}390$/$1{,}357$} \\
\addlinespace[2pt]
\multicolumn{2}{@{}l}{$S_{\mathrm{FROM\_J}} - S_{\mathrm{SELF}}$} & \ci{$+0.006639$}{$[0.003187,\,0.010175]$} & \ci{$+0.001306$}{$[0.000992,\,0.001647]$} & \ci{$+0.001122$}{$[0.000848,\,0.001424]$} & $-1$ / $+7$ items \\
\addlinespace[2pt]
\multicolumn{2}{@{}l}{$J_{\mathrm{FROM\_S}} - J_{\mathrm{SELF}}$} & \ci{$+0.003992$}{$[0.000960,\,0.007017]$} & \ci{$+0.000782$}{$[0.000487,\,0.001073]$} & \ci{$+0.001113$}{$[0.000845,\,0.001403]$} & $+3$ / $-14$ items \\
\addlinespace[2pt]
\multicolumn{2}{@{}l}{Interaction $\mathcal{I}_{SJ}$} & \ci{$+0.010631$}{$[0.008917,\,0.012355]$} & \ci{$+0.002088$}{$[0.001976,\,0.002208]$} & \ci{$+0.002235$}{$[0.001932,\,0.002648]$} & $+0.11$ / $-0.38\text{ pp}$ \\
\bottomrule
\end{tabular}}
\end{table}

\begin{table}[!htbp]
\centering
\caption{Circular moving-block C4 bootstrap sensitivity ($10{,}000$ draws, seed $20261014$) across block segment lengths $L \in \{1,2,4,8,16\}$ on Llama-3.2-3B and Fresh Qwen3-1.7B.}
\label{tab:a_c4_block}
\setlength{\tabcolsep}{5pt}
\renewcommand{\arraystretch}{1.14}
\footnotesize
\fittable{%
\begin{tabular}{@{}rrrr@{}}
\toprule
Segment $L$ & Llama $S$: $S(a_J) - S(a_S)$ & Llama $J$: $J(a_S) - J(a_J)$ & Qwen Fresh $F$: $F(a_Q) - F(a_F)$ \\
\midrule
$1$ & $+0.001122\;[+0.000837,\,+0.001423]$ & $+0.001113\;[+0.000838,\,+0.001409]$ & $+0.056240\;[+0.054000,\,+0.058631]$ \\
$2$ & $+0.001122\;[+0.000818,\,+0.001465]$ & $+0.001113\;[+0.000835,\,+0.001427]$ & $+0.056240\;[+0.053530,\,+0.059362]$ \\
$4$ & $+0.001122\;[+0.000790,\,+0.001514]$ & $+0.001113\;[+0.000828,\,+0.001434]$ & $+0.056240\;[+0.053067,\,+0.060145]$ \\
$8$ & $+0.001122\;[+0.000773,\,+0.001523]$ & $+0.001113\;[+0.000823,\,+0.001440]$ & $+0.056240\;[+0.052739,\,+0.060715]$ \\
$16$ & $+0.001122\;[+0.000723,\,+0.001569]$ & $+0.001113\;[+0.000794,\,+0.001444]$ & $+0.056240\;[+0.052538,\,+0.060974]$ \\
\bottomrule
\end{tabular}}
\end{table}

\clearpage
\section{Archival 6-Decimal Ledger for Full-Scale Capacity Regimes}
\label{app:capacity_archival}

On the $50$-step $Q_0$ full-scale ($196$-site, $22{,}020{,}096$ scales) pilot, consumed \textsc{Fit} metrics for $\text{lr} \in \{0.001, 0.003, 0.01\}$ were:
\begin{itemize}[leftmargin=1.6em,itemsep=1pt,topsep=2pt]
  \item $\text{lr}=0.001$: weighted \textsc{Fit} $= 2.373175$ ($\text{ARC KL}=0.072195$, $\text{Wiki NLL}=1.948155$, $\text{C4 NLL}=2.700797$);
  \item $\text{lr}=0.003$: weighted \textsc{Fit} $= 2.138045$ ($\text{ARC KL}=0.075077$, $\text{Wiki NLL}=1.347439$, $\text{C4 NLL}=2.653997$);
  \item $\text{lr}=0.010$: weighted \textsc{Fit} $= \mathbf{2.103050}$ ($\text{ARC KL}=0.112122$, $\text{Wiki NLL}=\mathbf{1.283866}$, $\text{C4 NLL}=2.631470$).
\end{itemize}
Table~\ref{tab:a_capacity} archives the $6$-decimal validation means for all $12$ Llama capacity states together with the principal paired contrasts and four-term regime interactions reproduced in \texttt{RESULTS.json}.

\begin{table}[!htbp]
\centering
\caption{Complete $6$-decimal archival ledger for the $12$ Llama-3.2-3B capacity states, paired regime contrasts, and four-term interactions ($10{,}000$ bootstrap draws, seed $20261014$).}
\label{tab:a_capacity}
\setlength{\tabcolsep}{5pt}
\renewcommand{\arraystretch}{1.18}
\footnotesize
\fittable{%
\begin{tabular}{@{}lrrrr@{}}
\toprule
State / Contrast / Interaction & ARC KL & WikiText NLL & C4 NLL & PIQA Norm / Raw ($\Delta\text{pp}$ or count) \\
\midrule
$Q_{\mathrm{FOUR}}$ / $S_{\mathrm{FOUR}}$ / $J_{\mathrm{FOUR}}$ & \ci{$0.102966$ / $0.100022$}{$0.105952$ ($J$)} & \ci{$2.248734$ / $2.263367$}{$2.264572$ ($J$)} & \ci{$2.706076$ / $2.705577$}{$2.704759$ ($J$)} & \ci{$1363$/$1352$; $1371$/$1369$}{$1387$/$1371$ ($J$)} \\
\addlinespace[2pt]
$Q_{\mathrm{HIGH200}}$ / $S_{\mathrm{HIGH200}}$ / $J_{\mathrm{HIGH200}}$ & \ci{$0.128318$ / $0.208868$}{$0.189428$ ($J$)} & \ci{$2.970805$ / $3.062988$}{$3.088196$ ($J$)} & \ci{$2.829036$ / $2.821918$}{$2.869482$ ($J$)} & \ci{$1372$/$1371$; $1336$/$1337$}{$1340$/$1332$ ($J$)} \\
\addlinespace[2pt]
$Q_{50}$ / $S_{50}$ / $J_{50}$ ($196\text{-site}, \text{lr}=10^{-3}$) & \ci{$0.093305$ / $0.096638$}{$0.096351$ ($J$)} & \ci{$2.201708$ / $2.201347$}{$2.203427$ ($J$)} & \ci{$2.686826$ / $2.683562$}{$2.684065$ ($J$)} & \ci{$1373$/$1357$; $1387$/$1382$}{$1392$/$1383$ ($J$)} \\
\addlinespace[2pt]
$Q_{200}$ / $S_{200}$ / $J_{200}$ ($196\text{-site}, \text{lr}=10^{-3}$) & \ci{$0.081001$ / $0.076198$}{$0.079192$ ($J$)} & \ci{$2.221406$ / $2.229744$}{$2.234604$ ($J$)} & \ci{$2.653432$ / $2.653460$}{$2.653716$ ($J$)} & \ci{$1382$/$1377$; $1388$/$1383$}{$1389$/$1392$ ($J$)} \\
\midrule
$Q_{\mathrm{HIGH200}} - Q_{\mathrm{FOUR}}$ & \ci{$+0.025352$}{$[+0.010389,+0.040480]$} & \ci{$+0.722071$}{$[+0.677692,+0.759524]$} & \ci{$+0.122960$}{$[+0.090738,+0.164620]$} & $+0.49\;[-0.76,+1.80]$ / $+1.03$ \\
\addlinespace[2pt]
$S_{\mathrm{HIGH200}} - S_{\mathrm{FOUR}}$ & \ci{$+0.108846$}{$[+0.089734,+0.128872]$} & \ci{$+0.799621$}{$[+0.740580,+0.841214]$} & \ci{$+0.116341$}{$[+0.108950,+0.125037]$} & $-1.90\;[-3.32,-0.44]$ / $-1.74$ \\
\addlinespace[2pt]
$J_{\mathrm{HIGH200}} - J_{\mathrm{FOUR}}$ & \ci{$+0.083475$}{$[+0.065291,+0.102281]$} & \ci{$+0.823624$}{$[+0.762302,+0.868224]$} & \ci{$+0.164723$}{$[+0.158805,+0.170880]$} & $-2.56\;[-4.03,-1.09]$ / $-2.12$ \\
\addlinespace[2pt]
$S_{200} - S_{\mathrm{HIGH200}}$ & \ci{$-0.132670$}{$[-0.151766,-0.114937]$} & \ci{$-0.833244$}{$[-0.853295,-0.811842]$} & \ci{$-0.168458$}{$[-0.176831,-0.161384]$} & $+2.83\;[+1.41,+4.24]$ / $+2.50$ \\
\addlinespace[2pt]
$J_{200} - J_{\mathrm{HIGH200}}$ & \ci{$-0.110236$}{$[-0.128179,-0.093060]$} & \ci{$-0.853592$}{$[-0.877638,-0.827537]$} & \ci{$-0.215766$}{$[-0.221940,-0.209841]$} & $+2.67\;[+1.20,+4.13]$ / $+3.26$ \\
\addlinespace[2pt]
$S_{200} - S_{\mathrm{FOUR}}$ & \ci{$-0.023823$}{$[-0.030619,-0.017305]$} & \ci{$-0.033623$}{$[-0.084585,-0.000713]$} & \ci{$-0.052117$}{$[-0.057055,-0.047759]$} & $+0.92\;[+0.11,+1.80]$ / $+0.76$ \\
\addlinespace[2pt]
$J_{200} - J_{\mathrm{FOUR}}$ & \ci{$-0.026761$}{$[-0.034082,-0.019760]$} & \ci{$-0.029968$}{$[-0.076151,+0.000739]$} & \ci{$-0.051043$}{$[-0.055874,-0.046771]$} & $+0.11\;[-0.71,+0.92]$ / $+1.14$ \\
\addlinespace[2pt]
$S_{200} - S_{50}$ & \ci{$-0.020440$}{$[-0.027052,-0.014102]$} & \ci{$+0.028397$}{$[+0.022517,+0.035429]$} & \ci{$-0.030101$}{$[-0.033118,-0.027253]$} & $+0.05\;[-0.49,+0.65]$ / $+0.05$ \\
\addlinespace[2pt]
$J_{200} - J_{50}$ & \ci{$-0.017159$}{$[-0.023920,-0.010562]$} & \ci{$+0.031176$}{$[+0.023012,+0.041919]$} & \ci{$-0.030350$}{$[-0.033523,-0.027364]$} & $-0.16\;[-0.82,+0.49]$ / $+0.49$ \\
\addlinespace[2pt]
$J_{200} - S_{200}$ & \ci{$+0.002993$}{$[+0.000094,+0.005986]$} & \ci{$+0.004860$}{$[+0.001375,+0.010689]$} & \ci{$+0.000255$}{$[-0.000415,+0.000921]$} & $+0.05\;[-0.71,+0.82]$ / $+0.49$ \\
\midrule
Int.\ \texttt{high\_S-Q} ($196\text{-}4$) & \ci{$+0.083494$}{$[+0.063195,+0.104538]$} & \ci{$+0.077550$}{$[+0.047202,+0.105908]$} & \ci{$-0.006619$}{$[-0.039815,+0.018508]$} & $-2.39\;[-4.03,-0.71]$ / $-2.77$ \\
\addlinespace[2pt]
Int.\ \texttt{high\_J-Q} ($196\text{-}4$) & \ci{$+0.058123$}{$[+0.039403,+0.076938]$} & \ci{$+0.101552$}{$[+0.068089,+0.132245]$} & \ci{$+0.041763$}{$[+0.001357,+0.072572]$} & $-3.05\;[-4.68,-1.36]$ / $-3.16$ \\
\addlinespace[2pt]
Int.\ \texttt{lr\_S-Q} ($10^{-3}\text{-}10^{-2}$) & \ci{$-0.085353$}{$[-0.105286,-0.065889]$} & \ci{$-0.083845$}{$[-0.106768,-0.056539]$} & \ci{$+0.007147$}{$[-0.017867,+0.040145]$} & $+2.29\;[+0.65,+3.92]$ / $+2.18$ \\
\addlinespace[2pt]
Int.\ \texttt{lr\_J-Q} ($10^{-3}\text{-}10^{-2}$) & \ci{$-0.062919$}{$[-0.080800,-0.044765]$} & \ci{$-0.104194$}{$[-0.132768,-0.070913]$} & \ci{$-0.040162$}{$[-0.070933,-0.000026]$} & $+2.12\;[+0.49,+3.75]$ / $+2.94$ \\
\bottomrule
\end{tabular}}
\end{table}

\clearpage
\section{Archival 6-Decimal Ledgers for the 24-Endpoint ARC Readout Audit}
\label{app:readout_archival}

Tables~\ref{tab:a_readout} and~\ref{tab:a_interaction} provide complete $6$-decimal archival records for all $24$ Qwen3-1.7B ARC readout endpoints ($6$ checkpoints $\times$ $4$ interfaces across $569$ items), the $5$-fold cross-fitted four-term interaction hierarchy on the $546$ standard \texttt{ABCD} items, and the post-hoc bias-subtraction and same-token ($\text{Chat}-\text{Bare}$) diagnostics.

\begin{table}[!htbp]
\centering
\caption{Complete $24$-endpoint archival $6$-decimal ledger for the Qwen3-1.7B ARC readout audit ($569$ items; $10{,}000$ joint bootstrap draws, seed $20261016$).}
\label{tab:a_readout}
\setlength{\tabcolsep}{4.8pt}
\renewcommand{\arraystretch}{1.14}
\footnotesize
\fittable{%
\begin{tabular}{@{}llrrrr@{}}
\toprule
Interface & State / Contrast & $D_{\mathrm{KL}}(p_{\mathrm{FP0}}\|p_\theta)$ $[95\%\text{ CI}]$ & Top-$1$ Agr.\ (\%) & Label Mass & Mean $\log$ Mass \\
\midrule
\multirow{4}{*}{\textsc{Legacy\_Bare}} & $\mathrm{FP}_0$ / $F_{200}$ / $F_{600}$ & $0$ / $0.116108$ / $0.118157$ & $100.0$ / $93.32$ / $94.02$ & $2.20\text{e-}9$ / $6.58\text{e-}5$ / $3.75\text{e-}5$ & $-26.52$ / $-11.37$ / $-11.84$ \\
 & $Q_0$ / $Q_{200}$ / $Q_{600}$ & $1.737541$ / $0.655221$ / $0.705274$ & $56.77$ / $77.15$ / $75.57$ & $1.49\text{e-}6$ / $2.35\text{e-}4$ / $2.32\text{e-}4$ & $-16.57$ / $-9.44$ / $-9.58$ \\
 & $Q_{600} - Q_{200}$ & $+0.050054\;[+0.017590,\,+0.083242]$ & $-1.58\;[-3.34,+0.18]$ & --- & --- \\
 & $F_{600} - F_{200}$ & $+0.002049\;[-0.005423,\,+0.009903]$ & $+0.70\;[-0.35,+1.76]$ & --- & --- \\
\midrule
\multirow{4}{*}{\textsc{Space\_Label}} & $\mathrm{FP}_0$ / $F_{200}$ / $F_{600}$ & $0$ / $0.095131$ / $0.095208$ & $100.0$ / $95.61$ / $95.78$ & $0.999737$ / $0.954430$ / $0.967801$ & $-0.0003$ / $-0.051$ / $-0.035$ \\
 & $Q_0$ / $Q_{200}$ / $Q_{600}$ & $1.203886$ / $0.653869$ / $0.702377$ & $65.20$ / $76.63$ / $77.15$ & $0.887008$ / $0.913305$ / $0.950837$ & $-0.141$ / $-0.098$ / $-0.054$ \\
 & $Q_{600} - Q_{200}$ & $+0.048508\;[+0.014776,\,+0.083622]$ & $+0.53\;[-1.23,+2.28]$ & Diff-in-diff vs.\ Bare: & $-0.001546$ \\
 & $F_{600} - F_{200}$ & $+0.000077\;[-0.006545,\,+0.006415]$ & $+0.18\;[-0.70,+1.05]$ & Diff-in-diff vs.\ Bare: & $-0.001972$ \\
\midrule
\multirow{4}{*}{\textsc{Answer\_Text}} & $\mathrm{FP}_0$ / $F_{200}$ / $F_{600}$ & $0$ / $0.549115$ / $0.565666$ & $100.0$ / $69.77$ / $69.24$ & n/a (multi-token) & n/a \\
 & $Q_0$ / $Q_{200}$ / $Q_{600}$ & $1.519828$ / $1.215586$ / $1.297265$ & $50.26$ / $52.90$ / $53.78$ & n/a (multi-token) & n/a \\
 & $Q_{600} - Q_{200}$ & $+0.081679\;[+0.051918,\,+0.110187]$ & $+0.88\;[-1.05,+2.81]$ & Diff-in-diff vs.\ Bare: & $+0.031625$ \\
 & $F_{600} - F_{200}$ & $+0.016551\;[+0.004496,\,+0.028952]$ & $-0.53\;[-2.11,+0.88]$ & Diff-in-diff vs.\ Bare: & $+0.014502$ \\
\midrule
\multirow{4}{*}{\textsc{Chat\_Label}} & $\mathrm{FP}_0$ / $F_{200}$ / $F_{600}$ & $0$ / $0.226612$ / $0.303319$ & $100.0$ / $95.78$ / $94.55$ & $0.960771$ / $0.957774$ / $0.973353$ & $-0.046$ / $-0.050$ / $-0.031$ \\
 & $Q_0$ / $Q_{200}$ / $Q_{600}$ & $5.600584$ / $4.336386$ / $4.048913$ & $29.70$ / $33.74$ / $34.80$ & $0.054343$ / $0.055684$ / $0.085244$ & $-5.80$ / $-4.65$ / $-3.98$ \\
 & $Q_{600} - Q_{200}$ & $-0.287473\;[-0.349671,\,-0.226663]$ & $+1.05\;[-0.18,+2.28]$ & Diff-in-diff vs.\ Bare: & $-0.337527$ \\
 & $F_{600} - F_{200}$ & $+0.076707\;[+0.031259,\,+0.127452]$ & $-1.23\;[-2.46,+0.00]$ & Diff-in-diff vs.\ Bare: & $+0.074658$ \\
\bottomrule
\end{tabular}}
\end{table}

\begin{table}[!htbp]
\centering
\caption{Part A: $5$-fold cross-fitted predictor hierarchy for the four-term relative $\text{Chat}-\text{Space}$ interaction $h_M = u_M - u_{\mathrm{FP0}}$ on $546$ \texttt{ABCD} items (seed $20261017$), alongside mean centered direct student shifts $u_M$. Part B: Post-hoc CPU bias-subtraction KL and teacher-agreement ledger (subtracting out-of-fold direct shift bias vs.\ excess-over-$\mathrm{FP}_0$ shift bias) and same-token $\text{Chat}-\text{Bare}$ static bias SSE reduction.}
\label{tab:a_interaction}
\setlength{\tabcolsep}{4.6pt}
\renewcommand{\arraystretch}{1.13}
\footnotesize
\fittable{%
\begin{tabular}{@{}lrrrrrr@{}}
\toprule
\multicolumn{7}{@{}l}{\emph{Part A: Four-Term Relative Interaction $h_M = u_M - u_{\mathrm{FP0}}$ Zero-Baseline SSE Reduction~\eqref{eq:sse_red} ($546$ \texttt{ABCD} items)}} \\
\midrule
State & Zero-Int.\ SSE & Bias SSE Red. & 1-Sc.$+$Bias Red. & 2-Sc.$+$Bias Red. & Chat Choice ($0 \to \text{2-Sc.}$) & Mean Direct Shift $u_M$ $(\texttt{A},\texttt{B},\texttt{C},\texttt{D})$ \\
\midrule
$Q_0$ & $121{,}726.83$ & $23.51\%$ & $88.91\%$ & $89.83\%$ & $169 \to 507/546$ & $(+6.584,\,-5.039,\,+0.062,\,-1.607)$ \\
$Q_{200}$ & $100{,}826.51$ & $18.32\%$ & $92.86\%$ & $93.71\%$ & $187 \to 504/546$ & $(+5.464,\,-3.927,\,+0.212,\,-1.749)$ \\
$Q_{600}$ [$Q_{\mathrm{SELF}}$] & $107{,}002.46$ & $17.88\%$ & $92.00\%$ & $93.20\%$ & $189 \to 494/546$ & $(+5.686,\,-3.766,\,-0.405,\,-1.515)$ \\
$F_{200}$ & $11{,}426.60$ & $3.58\%$ & $3.58\%$ & $73.35\%$ & $521 \to 527/546$ & $(-0.144,\,-0.875,\,+0.738,\,+0.281)$ \\
$F_{600}$ [$\mathrm{FP}_{\mathrm{SELF}}$] & $11{,}852.12$ & $8.78\%$ & $8.65\%$ & $70.28\%$ & $516 \to 524/546$ & $(-0.580,\,-0.683,\,+0.825,\,+0.438)$ \\
\midrule
\multicolumn{7}{@{}l}{\emph{Part B: Post-Hoc $5$-Fold Bias Subtraction on \textsc{Chat\_Label} \& Same-Token $\text{Chat}-\text{Bare}$ Control ($546$ \texttt{ABCD} items)}} \\
\midrule
State & Orig.\ KL & Direct-Debias KL & Excess-Debias KL & Orig.\ Agr. & Direct / Excess Agr. & $\text{Chat}-\text{Bare}$ Bias SSE ($\texttt{A}-\texttt{B}$) \\
\midrule
$Q_0$ & $5.579370$ & $2.059923$ & $1.924016$ & $30.04\%$ ($164$) & $36.08\%$ ($197$) / $38.64\%$ ($211$) & $82.75\%$ ($+15.315$) \\
$Q_{200}$ & $4.418689$ & $1.242016$ & $1.404370$ & $33.33\%$ ($182$) & $56.59\%$ ($309$) / $54.95\%$ ($300$) & $79.89\%$ ($+10.176$) \\
$Q_{600}$ [$Q_{\mathrm{SELF}}$] & $4.146755$ & $1.181062$ & $1.301902$ & $33.88\%$ ($185$) & $57.51\%$ ($314$) / $54.95\%$ ($300$) & $76.82\%$ ($+10.349$) \\
$F_{200}$ & $0.236110$ & $0.239778$ & $0.213062$ & $95.60\%$ ($522$) & $95.24\%$ ($520$) / $95.42\%$ ($521$) & $2.23\%$ ($+1.899$) \\
$F_{600}$ [$\mathrm{FP}_{\mathrm{SELF}}$] & $0.315838$ & $0.296039$ & $0.265576$ & $94.32\%$ ($515$) & $94.51\%$ ($516$) / $94.69\%$ ($517$) & $1.83\%$ ($+1.245$) \\
\bottomrule
\end{tabular}}
\end{table}

\end{document}